\documentclass[final,journal,twocolumn]{IEEEtran}
\usepackage{amssymb}
\usepackage{booktabs}
\usepackage[T1]{fontenc}
\newcommand{\coloneqq}{\mathrel{\mathop:}=}

\ifCLASSINFOpdf
  \usepackage[pdftex]{graphicx}
   \DeclareGraphicsExtensions{.pdf,.jpeg,.png}
\else
   \usepackage[dvips]{graphicx}
\fi

\usepackage{tikz,caption}
\usepackage{pgfplots}
\pgfplotsset{compat=newest}
\pgfplotsset{plot coordinates/math parser=false}
\newlength\figureheight
\newlength\figurewidth 
\usepackage{pgfplots}
\usepackage[cmex10]{amsmath}
\DeclareSymbolFont{AMSb}{U}{msb}{m}{n}
\DeclareMathSymbol{\C}{\mathalpha}{AMSb}{"43}
\DeclareMathSymbol{\R}{\mathalpha}{AMSb}{"52}
\usepackage{algorithmic}
\usepackage{array}
\usepackage{mdwmath}
\usepackage{mdwtab}
\usepackage{eqparbox}
\usepackage[caption=false,font=footnotesize]{subfig}
\usepackage{stfloats}
\ifCLASSOPTIONcaptionsoff
 \usepackage[nomarkers]{endfloat}
 \let\MYoriglatexcaption\caption
 \renewcommand{\caption}[2][\relax]{\MYoriglatexcaption[#2]{#2}}
\fi
\usepackage{url}
\usepackage{color}
\usetikzlibrary{plotmarks} 
\usetikzlibrary{external}
\usepgfplotslibrary{external} 
\tikzset{external/system call={pdflatex \tikzexternalcheckshellescape -shell-escape -halt-on-error -interaction=batchmode -jobname "\image" "\texsource"}}

\newcommand{\A}{\mathcal{A}}

\newcommand{\eps}{\varepsilon}

\def\cg{{{\sc cg}}}

\def\abs#1{\left|#1\right|} 

\begin{document}
%
\title{Generalizing diffuse interface methods on graphs: non-smooth potentials and hypergraphs}

%
%
%

\author{Jessica Bosch, Steffen Klamt, 
        and~Martin~Stoll
\thanks{
M.~Stoll and S.~Klamt are with the Max Planck Institute for Dynamics of Complex Technical Systems, Sandtorstr. 1, 39106
Magdeburg, Germany (stollm@mpi-magdeburg.mpg.de, klamt@mpi-magdeburg.mpg.de).
J.~Bosch is with the Department of Computer Science, The University of British Columbia, 201-2366 Main Mall, 
Vancouver, BC V6T 1Z4, Canada (jbosch@cs.ubc.ca).
}
}

\maketitle

\begin{abstract}
Diffuse interface methods have recently been introduced for the task of semi-supervised learning. The underlying model is well-known in materials science but was extended to graphs using a Ginzburg--Landau functional and the graph Laplacian. We here generalize the previously proposed model by a non-smooth potential function. Additionally, we show that the diffuse interface method can be used for the segmentation of data coming from hypergraphs. For this we show that the graph Laplacian in almost all cases is derived from hypergraph information. Additionally, we show that the formerly introduced hypergraph Laplacian coming from a relaxed optimization problem is well suited to be used within the diffuse interface method. We present computational experiments for graph and hypergraph Laplacians.
\end{abstract}

%
\IEEEpeerreviewmaketitle
\section{Introduction}
The classification of high-dimensional data on graphs is a challenging problem in many application areas \cite{ZhoS04,ShuNFOV13} and several techniques have been developed to efficiently tackle this problem. Recently, Bertozzi and Flenner \cite{BerF12} have established a method on the interface of graph based methods and partial differential equations (PDEs). Their method, which has already been extended to other cases (see \cite{GarMBFP14,MerGBFP14}), utilizes the information of the underlying graph via its graph Laplacian and then uses diffuse interface techniques for the separation of the given data into two classes. Diffuse interface techniques are a classical tool within the materials science community \cite{TC94,AllC79}. The new technique of Bertozzi and Flenner uses an approach taken from image inpainting based on phase-field methods \cite{BerEG07} for a semi-supervised learning problem. The use of phase-field models in image processing has seen many contributions (cf. \cite{esedog2006threshold,
chan2001active}).
 
To further use the inpainting analogy in the semi-supervised learning problem, the known or sampled data, which are used to train the method, can be considered the intact part of the image and we aim to restore the damaged or unknown part of the image. This works for both segmentation into two classes for binary images or into multiple classes for gray-valued or color images. PDE-based inpainting has been very successful \cite{ChaS01} and the technique introduced in \cite{BerF12} showed very promising results when compared to other methods such as the 1-Laplacian inverse power method (IPM) of Hein and B\"uhler \cite{hein2010inverse}. 

Our goal in this paper is the extension of the diffuse interface technique from using smooth potentials to the case of non-smooth potentials as well as the introduction of the diffuse interface approach when applied to hypergraph based segmentation. Non-smooth potentials are now widely used in many materials science applications \cite{EllG96,BlaBGSS12} and our previous work \cite{BosKSW14,BosS15} in image processing illustrated their importance also for image inpainting. The incorporation of these potentials into the graph based approach requires the use of an additional non-linear solver for which we propose a semi-smooth Newton method \cite{HIK02}. Furthermore, we show that the segmentation is not limited to two classes but extend this to the multiclass segmentation problem as considered in \cite{GarMBFP14}. Additionally, we aim at showing that the approach from \cite{BerF12} is so general that the underlying structural information do not necessarily have to come from the graph Laplacian but that the 
often very natural hypergraph formulation is well-suited for the combination with phase-field approaches both with smooth and non-smooth potentials. 

We start our discussion by introducing the graph Laplacian and the computation of some of its smallest eigenvalues. We then introduce the diffuse interface technique introduced in \cite{BerF12} and extend it to the case
when a non-smooth potential is used. This is done both for the two-classes segmentation problem as well as the multiclass segmentation. We further extend the existing approaches by illustrating
the applicability of diffuse interface methods on hypergraphs. Numerical results illustrate that the proposed methods work well on many test problems.
\section{The graph Laplacian and fundamentals}
We here consider an undirected graph $G=(V,E)$ consisting of a vertex set $V=\left\lbrace x_i\right\rbrace_{i=1}^{n}$ and the edge set $E$ \cite{Chu97}. Each edge $e\in E$ is a pair of nodes $(x_i,x_j)$ with $x_i\neq x_j$ and $x_i,x_j\in V$. For a weighted graph we also have a weight function $w:V\times V\rightarrow\mathbb{R}$ with $w(x_i,x_j)=w(x_j,x_i)\textrm{ for all } i,j$ for an undirected graph. We assume further that the function is positive for existing edges and zero otherwise. The degree of the vertex $x_i\in V$ is defined as 
$$
d(x_i)=\sum_{x_j\in V}w(x_i,x_j).
$$
The diagonal degree matrix $D\in\R^{n,n}$ is defined as $D_{i,i}=d(x_i).$
Now the crucial tool for further investigations is the graph Laplacian $L$ which is defined via 
$$
L(x_i,x_j)=
\begin{cases}
d(x_i)&\textnormal{ if }x_i=x_j  \\ 
-w(x_i,x_j) & \textnormal{ otherwise. }
\end{cases}
$$
It is clear that we can write $L=D-W$ with the entries of the weight matrix $W_{ij}$ given by $w(x_i,x_j).$ The Laplacian in this form is rarely used as typically its normalized form \cite{VLu07} is employed for segmentation purposes. The normalized Laplacian is defined by
$$
L_s=D^{-1/2}LD^{-1/2}=I-D^{-1/2}WD^{-1/2},
$$
which is a symmetric matrix. In contrast another normalized Laplacian of nonsymmetric form is given by
$$
L_w=D^{-1}L=I-D^{-1}W.
$$
We will use the eigenvalues of the symmetric and normalized graph Laplacian for numerical purposes later. We now discuss possibilities to compute the eigenvalues and eigenvectors of the matrix $L_s$.

\subsection{Computing eigenvalues of the Laplacian}
In practice the computation of several small eigenvalues of a matrix is a very challenging task. For small to moderate sizes the QR algorithm \cite{book::golubvanloan} is the method of choice for the computation of all eigenvalues of a matrix. For the computation of a subset of the eigenvalues the Lanczos algorithm for symmetric matrices and the Arnoldi algorithm for nonsymmetric matrices are typically chosen if the matrix is large and sparse \cite{lehoucq97arpack}. For large and sparse graphs the Laplacian will also be large and sparse and the authors in \cite{BerF12,GarMBFP14} suggest a Rayleigh-Chebyshev procedure \cite{And10}. Since the matrix is semi-definite a straightforward inverse iteration cannot be employed. One could consider projection techniques \cite{BL05} and employing suitable preconditioners is possible \cite{FreS08}. Our goal (cf. \cite{MerKB13}) is to compute the $k$ smallest eigenvalues of $L_s=D^{-1/2}LD^{-1/2}=I-D^{-1/2}WD^{-1/2}.$ 
For this it is clear that one could also focus on the largest eigenvalues $\lambda_j$ of the matrix $D^{-1/2}WD^{-1/2}$ as the eigenvalues of $L_s$ are given via $1-\lambda_j.$
One could use the Lanczos method for the computation of the largest eigenvalues as this method only requires the multiplication by the matrices $D^{-1/2}$ and $W$. 
Here the dominating cost is given by the application of $W,$ which potentially could be a dense matrix depending on the structure of the graph. To avoid such an expensive step a more advanced method was proposed by Bertozzi and Flenner in \cite{BerF12}. They introduce the well-known Nystr\"om extension \cite{FowBCM04,FowBM01}, which can work with submatrices of $L_s$ that are of much smaller dimension. The method operates by approximating the eigenpairs of $D^{-1/2}WD^{-1/2}$ using a quadrature rule with randomly chosen interpolation points. For simplicity, we will rely on the Lanczos process for $D^{-1/2}WD^{-1/2}$ via MATLAB's \texttt{eigs} function but recommend the use of randomized methods, such as the Nystr\"om extension, for large-scale graph segmentation.

\subsection{Weight function}
The choice of weight functions $w(x_i,x_j)$ is a crucial ingredient in the construction of the graph Laplacian. This choice will influence the performance of the segmentation process and the speed of the algorithm. This means that different choices of $w$ result in different segmentation results. The graph Laplacian will be crucially influenced by the weight matrix $W$. For example a sparse matrix $W$ will allow a much easier computation of the eigenvalues of $L_s.$ This means for complete graphs that the weight matrix needs to neglect certain relations between nodes whereas sparse graphs automatically result in sparse weight matrices. 

Typical choices for $w(x_i,x_j)$ are the Gaussian function
\begin{equation}
\label{weightgraph}
w(x_i,x_j)=\exp\left(-\frac{\mathrm{dist}(x_i,x_j)^2}{\sigma}\right)
\end{equation}
for some scaling parameter $\sigma$ and different choices for the metric $\mathrm{dist}(x_i,x_j)$ result in different methods. Another popular choice was introduced by Zelnik-Manor and Perona \cite{ZelP04} as
\begin{equation}
\label{weightgraph_ZMP}
w(x_i,x_j)=\exp\left(-\frac{\mathrm{dist}(x_i,x_j)^2}{\sqrt{\tau(x_i)\tau(x_j)}}\right)
\end{equation}
where $\tau(x_i)=\mathrm{dist}(x_i,x_k)$ and $\tau(x_j)=\mathrm{dist}(x_j,x_k)$ are local scalings of the weight to the $R$-th nearest neighbour\footnote{For the MATLAB\copyright\ codes computing the graph Laplacian we refer to \url{http://www.vision.caltech.edu/lihi/Demos/SelfTuningClustering.html}.}. It is clear that for the application in image processing the distance $\mathrm{dist}(x,y)$ is the difference between intensities of the pixels $y$ and $x.$ For color images this will be the sum of distances within the different channels. For other applications, e.g. machine learning, $\mathrm{dist}(x_i,x_j)$ could measure the Euclidean distance between the corresponding feature vectors (cf. \cite{BerF12}) of $x_i$ and $x_j$.
We have now the ingredients to compute the graph Laplacian as well as approximating $k$ of its smallest eigenvalues and introduce the diffuse interface techniques next.
\section{Diffuse interface methods on graphs with non-smooth potentials}
\label{sec::nonsmooth}
\subsection{Diffuse interface methods}
Diffuse interface methods are a classical and versatile tool in the simulation of materials science problems such as solidification processes \cite{AllC79,CahH58}. They are an indispensable tool for the simulation of phase separation processes but have over time spread to various other application areas ranging from biomembrane simulation \cite{WanD08} to image inpainting \cite{BerEG07,BosKSW14}. 

As these methods describe the separation of a mixed medium into two or more phases this methodology was recently extended by Bertozzi and Flenner \cite{BerF12}. These techniques, which are typically formulated in an infinite-dimensional setting, are now used within a graph-based formulation. The derivation of classical models such as the Allen--Cahn \cite{AllC79} or Cahn--Hilliard equations \cite{CahH58} is typically obtained from a gradient flow of the Ginzburg--Landau energy
\begin{equation}
\label{GL_scalar}
E(u)=\int\frac{\eps}{2}\abs{\nabla u}^2dx+\int\frac{1}{\eps}\psi(u)dx
\end{equation}
where $u$ is the phase-field and $\eps$ the interface parameter, which is typically assumed to be small. The function $\psi(u)$ is a potential that forces the phase-field $u$ to take values at either $u\approx -1$ or $u\approx 1$. We come back to the discussion of the choice of potential as this is one of the contributions of this paper. The minimization of the energy $E(u)$ follows a gradient flow, i.e., 
$$
\partial_{t}{u}=-\mathrm{grad}(E(u)).
$$
Different choices for the gradient lead to different evolution equations for the phase $u$. We here point to the well-known Allen--Cahn equation written as
\begin{align}
\label{allencahn1}
\partial_{t}{u}=\eps\Delta u-\eps^{-1}\psi'(u)
\end{align}
with given initial condition $u_0$ and zero Neumann boundary conditions. Here, $\psi'(u)$ is the derivative of a smooth potential $\psi(u)$. In the case of a non-smooth potential we obtain a variational inequality. The Allen--Cahn equation has also been used very successfully in image inpainting \cite{dobrosotskaya2008wavelet,li2015fast}. For this purpose Equation \eqref{allencahn1} is modified
\begin{align}
\label{allencahn2}
\partial_{t}{u}=\eps\Delta u-\eps^{-1}\psi'(u)+\omega(x)(f-u)
\end{align}
where $\omega(x)$ is a parameter that is zero in the damaged image domain $D$ and typically a large constant $\omega_0$ in the intact parts $\Omega\backslash D$. Here $f$ is the given image that we do not want to change in the undamaged part. 
\subsection{Diffuse interface methods on graphs}
In a very similar way, Bertozzi and Flenner formulated the semi-supervised learning problem. Here $f$ represents the learned data that have to be maintained throughout the evolution process. We want to derive a model that separates the domain $\Omega$ into two parts, i.e., two phases. The formulation for arbitrary information is inherently different as the $\Omega$ describes a set of points that we want to segment into two categories. For this the infinite-dimensional problem \eqref{allencahn1} is now defined using the description of the underlying graph Laplacian $L_s$ to given
\begin{equation}
\label{gl1}
E_s(u)=\frac{\eps}{2} u\cdot L_su+\sum_{x\in V}\frac{1}{\eps}\psi(u(x))+F(f,u)
\end{equation}
with the energy contribution $F(f,u)$ describing the fidelity term that would lead to $\omega(x)(f-u)$ in the continuous Allen--Cahn equation. 
Here, $u\cdot L_su$ is defined via
$$
\frac{\eps}{2} u\cdot L_su=
\frac{\eps}{2} 
\sum_{x_i,x_j}\frac{w(x_i,x_j)(u(x_i)-u(x_j))^2}{d(x_i,x_j)}
$$
and if $u=1$ in the set $A$ and $u=-1$ in $\bar{A}$ one obtains
\begin{align*}
\frac{\eps}{2} u\cdot L_su&=
\frac{\eps}{2} 
\sum_{\substack{x_i\in A,x_j\in \bar{A}\\ \textrm{ or }x_i\in \bar{A},x_j\in {A}}}\frac{w(x_i,x_j)(u(x_i)-u(x_j))^2}{d(x_i,x_j)}\\
&=\frac{4\eps}{2} \sum_{\substack{x_i\in A,x_j\in \bar{A}\\ \textrm{ or }x_i\in \bar{A},x_j\in {A}}}\frac{w(x_i,x_j)}{d(x_i,x_j)}.
\end{align*}
This clearly indicates that $u\cdot L_su$ is minimal if the weights across the interface, i.e. in between values from $A$ and $\bar{A},$ are minimized. For a more detailed discussion of the comparison of diffuse interface methods to other segmentation methods such as graph cuts and nonlocal means we refer to \cite{BerF12}.

We are now ready to write down the corresponding Allen--Cahn equation for the graph Laplacian as
\begin{equation}
 \label{allencahn_gl1}
 u_t=-\eps L_su-\eps^{-1}\psi'(u)+\omega(x)(f-u)
\end{equation}
(see \cite{van2014mean,luo2016convergence} for details).
Before discussing the details of the discretization we introduce a convexity splitting scheme that has been used very effectively for Cahn--Hilliard and Allen--Cahn equations with fidelity terms (see \cite{BerEG07,SchB11,Eyr98,BosKSW14}). For this the energy is split as
$$
E(u)=E_1(u)-E_2(u)
$$
with 
$$
E_1(u)=\int\frac{\eps}{2}\abs{\nabla u}^2dx+\int\frac{c}{2}\abs{u}^2dx
$$
and
$$
E_2(u)=-\int\frac{1}{\eps}\psi(u)dx+\int\frac{c}{2}\abs{u}^2dx-\int\frac{\omega(x)}{2}(f-u)^{2}dx.
$$
Using an implicit Euler for $E_1$ and explicit treatment for $E_2$ for the temporal evolution results in
\begin{multline*}
\frac{u(x)-\bar{u}(x)}{\tau}-\eps\Delta u(x)+cu(x)\\
=-\frac{1}{\eps}\psi'(\bar{u}(x))+c\bar{u}(x)+\omega(x)(f-\bar{u}(x)).
\end{multline*}
Note we did not introduce an index for the temporal discretization but rather assume that all values $u(x)$ are evaluated at the new time-point whereas $\bar{u}$ indicates the previous time-point. These equations are a model based on the infinite-dimensional formulation but our goal is to use the graph Laplacian based formulation as introduced in \cite{BerF12}. We obtain the same equations when our formulation is based on the graph Ginzburg--Landau energy, i.e.,
\begin{multline*}
\frac{u(x)-\bar{u}(x)}{\tau}+\eps L_s u(x)+cu(x)\\
=-\frac{1}{\eps}\psi'(\bar{u}(x))+c\bar{u}(x)+\omega(x)(f-\bar{u}(x)),
\end{multline*}
where the dimensionality of $u$ is adjusted to the size of the graph Laplacian. Assuming that $(\lambda_j,\phi_j)$ are the eigenpairs of $L_s$ we can write $u(x)=\sum_{k=1}^{m}\mathbf{u}_k\phi_k$
and from this we get
\begin{align}
&\frac{\mathbf{u}_k-\bar{\mathbf{u}}_k}{\tau}+\eps \lambda_k\mathbf{u}_k+c\mathbf{u}_k=-\frac{1}{\eps}\bar{\mathbf{b}}_k+c\bar{\mathbf{u}}_k+\bar{\mathbf{d}}_k
\end{align}
where 
$\bar{\mathbf{b}}=\psi'\left(\sum_{k=1}^{m}\bar{\mathbf{u}}_k\phi_k\right)$ and $\bar{\mathbf{d}}=\omega\left(f-\sum_{k=1}^{m}\bar{\mathbf{u}}_k\phi_k\right).$ We further rewrite this to obtain
\begin{align}
&\left(1+\eps\tau \lambda_k+c\tau\right)\mathbf{u}_k=\frac{\tau}{\eps}\bar{\mathbf{b}}_k+(1+c\tau)\bar{\mathbf{u}}_k+\tau\bar{\mathbf{d}}_k.
\end{align}
With the choice of $\psi(u)=\frac{1}{4}(u^2-1)^2$ we obtain the scheme introduced in \cite{BerF12}. 
\subsection{Non-smooth potentials}
In classical phase-field simulations the choice of potential function typically plays a crucial role and non-smooth potentials have proven to allow for the most realistic reproductions of processes in materials science. For this the well-known obstacle potential can be used. In more detail, we consider
\begin{equation}
\label{obstacle_pot}
\psi_{ns}(u)\coloneqq \left\{\begin{array}{rl}\frac{1}{2}(1-u^2), & -1\leq u \leq 1\\ \infty, & \textup{otherwise},\end{array}\right.
\end{equation}
and obtain the following modified Allen--Cahn equation
\begin{align}
 \partial_{t}{u}&=\varepsilon\Delta{u}-\frac{1}{\varepsilon}(\psi_{0}'{(u)}+\mu)+\omega(x)(f-u),\label{CH1_mod2}\\
\mu&\in \partial{\beta_{[-1,1]}(u)},\label{CH2_mod2}\\
-1\leq u &\leq 1,\label{CH3_mod2}\\
\frac{\partial{u}}{\partial{n}}=\frac{\partial{\Delta{u}}}{\partial{n}}&=0\quad\textup{on}\ \partial\Omega.\label{CH4_mod2}
\end{align} 
Here we have written $\psi_{ns}$ in (\ref{obstacle_pot}) via the indicator function 
as
\begin{displaymath}
\psi_{ns}{(u)}=\psi_{0}{(u)}+\mathbf{\mathit{I}}_{[-1,1]}{(u)}
\end{displaymath}
and $\psi_{0}(u):=\frac{1}{2}(1-u^2)$.
We now follow a well-known approach by regularizing the energy with the Moreau-Yosida penalty term \cite{BosS15,HinHT11} and obtain
\begin{multline*}	
E(u_{\nu})=\int_{\Omega}\frac{\varepsilon}{2}|\nabla{u_{\nu}}|^{2}+\frac{1}{\varepsilon}\psi_{0}{(u_{\nu})}\\
+\frac{1}{2\nu}|\max(0,u_{\nu}-1)|^{2}+\frac{1}{2\nu}|\min(0,u_{\nu}+1)|^{2}dx,
\end{multline*}
with $\nu$ the penalty parameter. Again, we consider the convexity splitting for this energy and obtain
\begin{multline*}	
E_1(u_{\nu})=\int\frac{\eps}{2}\abs{\nabla u_{\nu}}^2dx+\int\frac{c}{2}\abs{u_{\nu}}^2dx\\
+\int\frac{1}{2\nu}|\max(0,u_{\nu}-1)|^{2}+\frac{1}{2\nu}|\min(0,u_{\nu})|^{2}dx
\end{multline*}
and
\begin{multline*}
E_2(u_{\nu})=-\int\frac{1}{\eps}\psi_{0}(u_{\nu})dx\\
+\int\frac{c}{2}\abs{u_{\nu}}^2dx-\int\frac{\omega(x)}{2}(f-u_{\nu})^{2}dx.
\end{multline*}
This leads to the following evolution equation
\begin{multline}
\label{nonlinear1}
\frac{u_{\nu}-\bar{u}_{\nu}}{\tau}-\eps\Delta u_{\nu}+cu_{\nu}+\theta_{\nu}(u_{\nu})\\
=-\frac{1}{\eps}\psi_0'(\bar u)+c \bar u+\omega(f-\bar u),
\end{multline}
where 
\begin{displaymath} 
\theta_{\nu}(u_{\nu})\coloneqq\frac{1}{\nu}\textup{max}(0,u_{\nu}-1)+\frac{1}{\nu}\textup{min}(0,u_{\nu}+1).
\end{displaymath}
In the previous setup the nonlinearity coming from the potential term was shifted towards the right-hand side as it was treated explicitly. In the non-smooth setting we obtain a nonlinear relation due to  the non-smooth relation given by $\theta_{\nu}(u_{\nu})$, which we treat with the well-known semi-smooth Newton method \cite{HIK02}. For \eqref{nonlinear1} written as 
\begin{align*}
F(u_\nu)&=(c+\frac{1}{\tau})u_{\nu}-\eps\Delta u_{\nu}+\theta_{\nu}(u_{\nu})+\frac{1}{\eps}\psi_0'(\bar u)\\
&\quad\ -(c+\frac{1}{\tau})\bar u-\omega(f-\bar u)\\
&=0
\end{align*}
the Newton system is given via
$$
u^{(l+1)}_\nu=u_{\nu}^{(l)}-
G(u_\nu^{(l)})^{-1}F(u_\nu^{(l)}).
$$
We define the sets
\begin{align*}
\mathcal{A}(u_{\nu})&:=\left\lbrace x\in\Omega : u_{\nu}>1\textnormal{ or } u_{\nu}<-1\right\rbrace,\\
\mathcal{A}_{+}(u_{\nu})&:=\left\lbrace x\in\Omega : u_{\nu}>1\right\rbrace,\\
\mathcal{A}_{-}(u_{\nu})&:=\left\lbrace x\in\Omega : u_{\nu}<-1\right\rbrace,\\
\end{align*}
and write down the Newton system as
\begin{multline*}
G(u_{\nu}^{(l)})u_{\nu}^{(l+1)}\\
\begin{aligned}
&=G(u_{\nu}^{(l)})u_{\nu}^{(l)}-F(u_\nu^{(l)})\\
&=-\nu^{-1}\left(\chi_{\A_{-}(u_{\nu}^{(l)})}-\chi_{\A_{+}(u_{\nu}^{(l)})}\right)\mathbf{1}+(\frac{1}{\eps}+c+\frac{1}{\tau}) \bar u\\
&\quad\ +\omega(f-\bar{u})
\end{aligned}
\end{multline*}
where $G(u_{\nu}^{(l)}):=(c+\frac{1}{\tau})I-\eps \Delta+\frac{1}{\nu}\chi_{\A(u_{\nu}^{(l)})}$ with $I$ the identity operator. Again, we have first introduced the classical problem. The equivalent formulation using the graph Laplacian
is given via
\begin{multline}
\label{nonlinear2}
\frac{u_{\nu}-\bar{u}_{\nu}}{\tau}+\eps L_s u_{\nu}+cu_{\nu}+\theta_{\nu}(u_{\nu})\\
=-\frac{1}{\eps}\psi_0'(\bar{u})+c\bar{u}+\omega(f-\bar{u}),
\end{multline}
and we obtain the Newton system 
\begin{multline*}
G(u_{\nu}^{(l)})u_{\nu}^{(l+1)}=-\nu^{-1}\left(\chi_{\A_{-}(u_{\nu}^{(l)})}-\chi_{\A_{+}(u_{\nu}^{(l)})}\right)\mathbf{1}\\+
(\frac{1}{\eps}+c+\frac{1}{\tau})\bar u+\omega(f-\bar{u})
\end{multline*}
where $G(u_{\nu}^{(l)}):=(c+\frac{1}{\tau})I+\eps L_s+\frac{1}{\nu}\chi_{\A(u_{\nu}^{(l)})}$ with $I$ the identity matrix. In the following we drop the index $\nu$. This  Newton system is the equivalent to the infinite dimensional Newton system and in a Galerkin fashion we assume 
$u^{(l)}=\sum_{k=1}^{m}\mathbf{u}_{k,l}\phi_k=\Phi\mathbf{u}_{(l)}$ with a small number $m$ of terms chosen as the projection basis. This results in the projected system
\begin{multline}
\label{NSsystem}
\Phi^{T}G(u_{}^{(l)})\Phi\mathbf{u}_{(l+1)}=-\frac{1}{\nu}\Phi^{T}\left(\chi_{\A_{-}(u^{(l)})}-\chi_{\A_{+}(u^{(l)})}\right)\mathbf{1}\\+
 (\frac{1}{\eps}+c+\frac{1}{\tau})\Phi^{T}\Phi\bar{\mathbf{u}}+\Phi^{T}\omega(f-\Phi\bar{\mathbf{u}}).
\end{multline}
Here the crucial operator becomes 
$$
\Phi^{T}G(u_{}^{(l)})\Phi=(c+\frac{1}{\tau})I+\eps \Lambda+\frac{1}{\nu}\Phi^{T}\chi_{\A(u^{(l)})}\Phi
$$
where $\Lambda$ is the diagonal matrix containing the $m$ eigenvalues used in the approximation.
It is clear that (\ref{NSsystem}) requires the solution of a small $m\times m$ linear system for which we use the \cg\ method \cite{cg} or use a direct solver based on a factorisation of the matrix.

\section{Diffuse interface methods on graphs -- The vector-valued case}
\subsection{Vector-valued smooth diffuse interface methods}
Section \ref{sec::nonsmooth} was devoted to scalar diffuse interface models on graphs. In this section, we present 
their generalization to the vector-valued case. This can then be used for the multiclass segmentation problem.\\

In practice, often more than two components occur; see, e.g., the biomembrane simulation \cite{WanD08}, image inpainting of gray value 
images \cite{BosS15}, as for this the diffuse interface models have been extended to deal with multi-component systems. 
The Ginzburg--Landau energy for two components in (\ref{GL_scalar}) generalizes to
\begin{equation}
\label{GL_vector}
E({\bf u})=\int\frac{\eps}{2}\sum_{i=1}^{K}{\abs{\nabla u_{i}}^2}dx+\int\frac{1}{\eps}\psi({\bf u})dx
\end{equation}
for $K>2$ components. Here, ${\bf u}=(u_{1},\ldots,u_{K})^{T}$ is now the vector-valued phase-field, and 
the potential function $\psi({\bf u})$ has $K$ distinct minima instead of two. 
This section deals with smooth potentials, and the smooth potential in the scalar case generalizes to the 
vector-valued case as $\psi({\bf u})=\frac{1}{4}\sum_{i=1}^{K}{u_{i}^{2}(1-u_{i})^{2}}$. 
We come back to the discussion of non-smooth potentials in Section \ref{subsec:vv:nonsmooth}.\\

Recently, Garcia-Cardona et al.~\cite{GarMBFP14} as well as Merkurjev et al.~\cite{MerGBFP14} have extended 
these continuous models to the graph domain. In the following, we summarize their approach. 
As before, $n$ is the number of data points. We introduce the matrix 
$U=({\bf u}_{1},\ldots,{\bf u}_{n})^{T}\in\mathbb{R}^{n, K}$. Here, the $k$th component of 
${\bf u}_{i}\in\mathbb{R}^{K}$ is the strength for data point $i$ to belong to class $k$. 
For each node $i$, the vector ${\bf u}_{i}$ has to be an element of the Gibbs simplex $\Sigma^{K}$
\begin{displaymath}
 \Sigma^{K}:=\left\{(x_{1},\ldots,x_{K})^{T}\in[0,1]^{K}\left\vert\, \sum_{k=1}^{K}{x_{k}}=1\right.\right\}.
\end{displaymath}

The Ginzburg--Landau energy functional on graphs in (\ref{gl1}) generalizes to the multiclass case as
\begin{equation}
\label{gl_vv}
E(U)=\frac{\eps}{2} \langle U, L_{s}U\rangle+\frac{1}{\epsilon}\psi(U)+F(\hat U,U).
\end{equation}
Here,
\begin{displaymath}
 \langle U, L_{s} U\rangle=\textup{trace}(U^{T}L_{s}U)
\end{displaymath}
measures variations in the vector field, the potential term
\begin{displaymath}
 \psi(U)=\frac{1}{2}\sum_{i\in V}{\left(\prod_{k=1}^{K}{\frac{1}{4}||{\bf u}_{i}-{\bf e}_{k}}||_{L_{1}}^{2}\right)}
\end{displaymath} 
drives the system closer to the pure phases, and the fidelity term
\begin{displaymath}
 F(\hat U,U)=\sum_{i\in V}{\frac{\omega}{2}||\hat{\bf u}_{i}-{\bf u}_{i}||^{2}_{L_{2}}}
\end{displaymath} 
enables the encoding of a priori information with $\hat U=(\hat{\bf u}_{1},\ldots,\hat{\bf u}_{n})^{T}$ representing the learned data. 
In the potential term, ${\bf e}_{k}\in\mathbb{R}^{K}$ is the vector whose $k$th component equals one and all other components
vanish. The vectors ${\bf e}_{1},\ldots,{\bf e}_{K}$ correspond to the pure phases. 
Note that the authors use an $L_{1}$-norm for the potential term as it prevents an undesirable minimum from occurring at the center of the 
simplex, as would be the case with an $L_{2}$-norm for large $K$.\\

As in Section \ref{sec::nonsmooth}, the authors use a convexity splitting scheme to minimize the Ginzburg--Landau functional in the 
phase-field approach. For this, the energy (\ref{gl_vv}) is split as
\begin{displaymath}
 E(U)=E_1(U)-E_2(U)
\end{displaymath}
with
\begin{displaymath}
 E_1(U)=\frac{\eps}{2}\langle U, L_{s}U\rangle+\frac{c}{2}\langle U,U\rangle
\end{displaymath}
and
\begin{displaymath}
E_2(U)=-\frac{1}{\epsilon}\psi(U)-F(\hat U,U)+\frac{c}{2}\langle U,U\rangle.
\end{displaymath}
In order to guarantee the convexity of the energy terms, we require $c\geq\omega+\frac{1}{\epsilon}$. The convexity splitting scheme results in
\begin{equation}
 \label{CS_vv}
\frac{U-\bar{U}}{\tau}+\eps L_{s}U+cU=-\frac{1}{2\eps}T(\bar{U})+c\bar{U}+\omega(\hat{U}-\bar{U}),
\end{equation}
where the elements $T_{ik}$ of the matrix $T(\bar{U})$ are given as
\begin{displaymath}
 T_{ik}=\sum_{l=1}^{K}{\frac{1}{2}\left(1-2\delta_{kl}\right)||\bar{\bf u}_{i}-{\bf e}_{l}||_{L_{1}}}\prod_{m=1,m\neq l}^{K}{\frac{1}{4}||\bar{\bf u}_{i}-{\bf e}_{m}}||_{L_{1}}^{2}.
\end{displaymath}
Again, we assume that all values $U$ are evaluated at the new time-point 
whereas $\bar{U}$ indicates the previous time-point. 
Multiplying (\ref{CS_vv}) by $\Phi^{T}$ from the left and 
using the eigendecomposition $L_s=\Phi\Lambda\Phi^{T}$, we obtain
\begin{equation}
 \label{CS_vv_2}
 \mathcal{U}=B^{-1}\left[(1+c\tau)\bar{\mathcal{U}}-\frac{\tau}{2\epsilon}\Phi^{T}T(\bar{U})+\tau\omega(\hat{\mathcal{U}}-\bar{\mathcal{U}})\right],
\end{equation}
where all calligraphic fonts have the meaning $\mathcal{U}=\Phi^{T}U$. 
Since $B=(1+c\tau)I+\epsilon\tau\Lambda$ is a diagonal matrix with positive entries, its inverse is easy to apply.\\

After the update, we have to project the solution back to the Gibbs simplex $\Sigma^{K}$. In order to do this, we use of the projection procedure in \cite{CheY11}. For the initialization of the segmentation problem, we first assign random values from the standard uniform distribution on $(0,1)$ to the nodes. 
Then, we project the result to the Gibbs simplex $\Sigma^{K}$ and set the values in the fidelity points to the pure phases. Here, we finish the presentation of the model proposed in \cite{GarMBFP14,MerGBFP14}. Next, we extend this approach to the use of non-smooth potentials.

\subsection{Vector-valued non-smooth diffuse interface methods}
\label{subsec:vv:nonsmooth}
In this section, we extend the approach above to the use of non-smooth potentials. 
We start with the continuous setting. 
The potential function in 
(\ref{GL_vector}) is now given as
\begin{equation}
  \label{vv_nonsmooth_pot}
  \psi({\bf u})= \left\{\begin{array}{rl}\psi_{0}({\bf u}) & {\bf u}\in\Sigma^{K},\\ \infty & \textup{otherwise},\end{array}\right.
\end{equation}
where the smooth part is given as $\psi_{0}({\bf u})=-\frac{1}{2}{\bf u}\cdot T{\bf u}$. 
Here, $T\in\mathbb{R}^{K, K}$ is a symmetric matrix, which contains constant interaction parameters $T_{ij}$. 
From physical considerations, $T$ must have at least one positive eigenvalue. 
A typical choice is $T = I-{\bf 1} {\bf 1}^{T}$ with ${\bf 1} = (1,\ldots, 1)^{T}\in\mathbb{R}^{K}$ and the identity 
matrix $I\in\mathbb{R}^{K, K}$, 
which means that the interaction between all different components is equal and no self-interaction occurs. 
In the numerical examples, we work with this choice of $T$.\\

As before in the scalar case, we propose to regularize the energy with a Moreau--Yosida 
penalty term and obtain
\begin{multline}
\label{GL_vector_MY}
E({\bf u}_{\nu})=\int\frac{\eps}{2}\sum_{i=1}^{K}{\abs{\nabla u_{\nu,i}}^2+\frac{1}{\eps}\psi_{0}({\bf u}_{\nu})}\\+\frac{1}{2\nu}\sum_{i=1}^{K}{\abs{\min(0,u_{\nu,i})}^{2}}dx.
\end{multline}
Here, $\nu$ is again the penalty parameter. Applying the convexity splitting scheme to (\ref{GL_vector_MY}) 
in the same way as in the non-smooth scalar case, we obtain the following time-discrete scheme
\begin{multline}
 \label{CS_vv_nonsmooth}
\frac{u_{\nu,i}-\bar{u}_{\nu,i}}{\tau}-\eps \Delta u_{\nu,i}+c u_{\nu,i}+\theta_{\nu}(u_{\nu,i})\\
=\frac{1}{\eps}(T\bar{\bf u})_{i}+c \bar{u}_{i}+\omega(\hat{u}_{i}-\bar{u}_{i})
\end{multline}
for $i=1,\ldots,K$, where
\begin{displaymath}
 \theta_{\nu}(u_{\nu,i}):=\frac{1}{\nu}\min(0,u_{\nu,i}).
\end{displaymath}
In order to guarantee the convexity of the energy terms, we require $c\geq\omega$.\\

Next, if we write (\ref{CS_vv_nonsmooth}) in the form $F_{i}(u_{\nu,i})=0$ for $i=1,\ldots,K$, the 
semi-smooth Newton system
\begin{displaymath}
 u_{\nu,i}^{(l+1)}=u_{\nu,i}^{(l)}-G_{i}(u_{\nu,i}^{(l)})^{-1}F_{i}(u_{\nu,i}^{(l)})
\end{displaymath}
is given as
\begin{displaymath}
 G_{i}(u_{\nu,i}^{(l)})u_{\nu,i}^{(l+1)}=(c+\frac{1}{\tau})\bar{u}_{i}+\frac{1}{\eps}(T\bar{\bf u})_{i}+\omega(\hat{u}_{i}-\bar{u}_{i}),
\end{displaymath}
where $G_{i}(u_{\nu,i}^{(l)}):=(c+\frac{1}{\tau})I-\eps\Delta+\frac{1}{\nu}\chi_{\mathcal{A}(u_{\nu,i}^{(l)})}$ 
with
\begin{displaymath}
 \mathcal{A}(u_{\nu,i}^{(l)}):=\{x\in\Omega\colon u_{\nu,i}^{(l)}(x)<0\}.
\end{displaymath}
This is the classical problem formulation. In the graph domain, (\ref{CS_vv_nonsmooth}) reads
\begin{multline}
 \label{CS_vv_nonsmooth_graph}
\frac{{\bf u}_{\nu,i}-\bar{{\bf u}}_{\nu,i}}{\tau}+\eps L_{s} {\bf u}_{\nu,i}+c {\bf u}_{\nu,i}+\theta_{\nu}({\bf u}_{\nu,i})\\
=\frac{1}{\eps}(T\bar{U}^{T})_{i}+c \bar{{\bf u}}_{i}+\omega(\hat{{\bf u}}_{i}-\bar{{\bf u}}_{i})
\end{multline}
for $i=1,\ldots,K$, where $\bar{U}=({\bf \bar{u}}_{1},\ldots,{\bf \bar{u}}_{K})\in\mathbb{R}^{n, K}$ similar to the previous section and 
\begin{displaymath}
 (T\bar{U}^{T})_{i}=-\sum_{j=1,j\neq i}^{K}{\bar{{\bf u}}_{j}}.
\end{displaymath}
The resulting Newton system is given as
\begin{equation}
\label{CS_vv_nonsmooth_trans}
 G_{i}({\bf u}_{\nu,i}^{(l)}){\bf u}_{\nu,i}^{(l+1)}=(c+\frac{1}{\tau})\bar{{\bf u}}_{i}+\frac{1}{\eps}(T\bar{U}^{T})_{i}+\omega(\hat{{\bf u}}_{i}-\bar{{\bf u}}_{i}),
\end{equation}
where $G_{i}({\bf u}_{\nu,i}^{(l)}):=(c+\frac{1}{\tau})I+\eps L_{s}+\frac{1}{\nu}\chi_{\mathcal{A}({\bf u}_{\nu,i}^{(l)})}$. 
Multiplying (\ref{CS_vv_nonsmooth_trans}) by $\Phi^{T}$ from the left and 
using the eigendecomposition $L_s=\Phi\Lambda\Phi^{T}$, we obtain
\begin{displaymath}
 G_{i}({\bf u}_{\nu,i}^{(l)}){\bf \mathcal{U}}_{\nu,i}^{(l+1)}=(c+\frac{1}{\tau})\bar{{\bf \mathcal{U}}}_{i}+\frac{1}{\eps}\Phi^{T}(T\bar{U}^{T})_{i}+\omega(\hat{{\bf \mathcal{U}}}_{i}-\bar{{\bf \mathcal{U}}}_{i}),
\end{displaymath}
where $G_{i}({\bf u}_{\nu,i}^{(l)}):=(c+\frac{1}{\tau})I+\eps \Lambda+\frac{1}{\nu}\Phi^{T}\chi_{\mathcal{A}({\bf u}_{\nu,i}^{(l)})}\Phi$ 
and all calligraphic fonts have the meaning $\mathcal{U}=\Phi^{T}{\bf u}$. Since this requires the solution of a small $m\times m$ linear system, 
we make use of MATLAB's backslash command.\\

Finally, after each time step, we project the solution back to the Gibbs simplex $\Sigma^{K}$ using the procedure in \cite{CheY11}.\\

Here, we finish the discussion about diffuse interface methods on graphs. 
Next, we introduce the diffuse interface approach when applied to hypergraph based segmentation.

\section{Hypergraphs and Laplacians}

In this section we want to show at how to generalize the before mentioned methodology to the case of hypergraphs.

A hypergraph is considered as $G=(V,E)$ with $V=\cup\left\lbrace x_i\right\rbrace$ a family of objects and $E$ a family of subsets $e$ of $V$ such that $\cup_{e\in E}=V.$ We call $V$ the vertices and $E$ the hyperedge set of $G$. If a weight $w(e)$ is associated with each hyperedge then the hypergraph is called weighted. We can also define the degree $d(x_j)$ as
$d(x_j)=\sum_{\left\lbrace e\in E:x_j\in e\right\rbrace}w(e).$ Also the edge in a hypergraph has a degree which is simply $\delta(e)=\abs{e}.$ The matrix $H\in\R^{\abs{V},\abs{E}}$ is the incidence matrix of the hypergraph where the rows correspond to the vertices and the columns to the hyperedges. In most applications, the entry $H_{i,j}$ is equal to one if the vertex $x_i$ is contained in the set that defines the hyperedge $j,$ otherwise the entries are set to zero.  
In all our applications the set of hyperedges refers to the different attributes that describe the problem.
The matrices $D_V$ and $D_E$ are diagonal matrices containing the degrees of the vertices and hyperedges, respectively. And the diagonal matrix $W_H$ is the weight matrix containing the weights of the hyperedges. One can then define the adjacency matrix $HW_HH^{T}-D_V.$ 

One might now wonder why the introduction of a hypergraph is a useful concept in the segmentation of data. Previous work explicitly using hypergraphs is given in \cite{ZhoHS06,hein2013total}. We here want to point out that in fact most real-world examples are initially represented via hypergraphs be it the image segmentation mentioned earlier where each vertex, i.e., pixel, has an associated vector of RGB values or the congress voting records used in \cite{BerF12} where for each congressman the voting record is stored in a feature vector. Since the incidence matrix of the hypergraph is naturally not square, in order to use the graph Laplacian the structure has to be transformed to a graph to represent pairwise relationships. 
To obtain pairwise relationships in both of these examples the computation of the square weight matrix $W$ from \eqref{weightgraph} and hence the computation of the distance between two feature vectors for example allows a one-to-one relation between the different vertices and hence the segmentation via the graph Laplacian. 

This means that in principal the methodology introduced earlier already takes hypergraph information that are then projected onto a simple graph where the information from the hyperedges is projected into the weight matrix $W$. 

We here present an alternative approach to project the hypergraph information onto a graph, i.e., to create pairwise relationships of hypergraph data. This approach is based on a relaxed problem that one considers instead of the NP hard cut problem. In more detail, one typically considers a relaxed optimization problem (cf. \cite{ZhoHS06})
\begin{align}
 \mathrm{argmin}_{u\in\R^{\abs{V}}}\frac{1}{2}\sum_{e\in E}\sum_{\left\lbrace x_i,x_j\right\rbrace\subseteq e }
 \frac{w(e)}{\delta(e)}\left(\frac{u(x_i)}{\sqrt{d(x_i)}}-\frac{u(x_j)}{\sqrt{d(x_j)}}\right)^2
\end{align}
subject to
\begin{align}
\sum_{x_j\in V}f(x_j)^2=1, \sum_{x_j\in V}u(x_j)\sqrt{d(x_j)}=0.
\end{align}
Defining the matrices $\Theta=D^{-1/2}_{V}HW_HD_{E}^{-1}H^{T}D_{V}^{-1/2}$ and $L_s=I-\Theta$ it was shown in \cite{ZhoHS06} that
\begin{align}
\label{hypeqi}
\sum_{e\in E}\sum_{\left\lbrace x_i,x_j\right\rbrace\subseteq e }
 \frac{w(e)}{\delta(e)}\left(\frac{u(x_i)}{\sqrt{d(x_i)}}-\frac{u(x_j)}{\sqrt{d(x_j)}}\right)^2=u^{T}L_su.
\end{align}
It is clear that due to $W$ and $D_E$ being diagonal matrices that the matrix $L_s$ is symmetric and the definiteness follows from \eqref{hypeqi}.
If $u=1$ in the set $A$ and $u=-1$ in $\bar{A}$ one obtains
\begin{align}
\label{hg1}u^{T}L_su=&\sum_{e\in E}\left(\sum_{\substack{x_i\in A,x_j\in\bar{A} \\\textrm{ or }\\x_i\in\bar{A},x_j\in A}}
 \frac{w(e)}{\delta(e)}\left(\frac{1}{\sqrt{d(x_i)}}+\frac{1}{\sqrt{d(x_j)}}\right)^2\right.\\
\label{hg2} +&\sum_{\substack{x_i\in A, \\x_j\in {A}}}
 \frac{w(e)}{\delta(e)}\left(\frac{1}{\sqrt{d(x_i)}}-\frac{1}{\sqrt{d(x_j)}}\right)^2\\
\label{hg3}  +&\sum_{\substack{x_i\in \bar{A}, \\x_j\in\bar{A}}}
 \left.\frac{w(e)}{\delta(e)}\left(\frac{-1}{\sqrt{d(x_i)}}+\frac{1}{\sqrt{d(x_j)}}\right)^2\right).
\end{align}
The last equation motivates the use of the diffuse interface approach as assuming that the degrees of the vertices are similar then \eqref{hg1} is the dominating term and hence minimization using the hypergraph Laplacian $u^{T}L_su$ is achieved if the weights across the interface are minimal.
We hence use the hypergraph Laplacian in the same way as the graph Laplacian for the segmentation of the vertices and run all diffuse interface models with the eigenvectors and eigenvalues of the hypergraph Laplacian instead of the graph Laplacian that could also be derived from hypergraph information. In the numerical experiments presented next we still use the original naming of the hypergraph and graph Laplacian even though both have been derived from hypergraph data.
\section{Numerical Experiments}
The aim is to show that the methods introduced in this paper are effective and we chose to compare the smooth and non-smooth potential version of the diffuse interface method for graph-based and hypergraph-based problems. Bertozzi and Flenner compare the diffuse interface approach in \cite{BerF12} to many other techniques such as the p-Laplacian \cite{szlam2009total} with favourable outcome for their approach. The computation of the eigenvalues is based on \texttt{eigs} from MATLAB, which uses the Lanczos process for $D^{-1/2}WD^{-1/2}$ in the graph case and for $\Theta$ in the hypergraph Laplacian. The results presented here are snapshots of a high-dimensional space of parameters that can be chosen and we want to illustrate the performance with respect to varying these parameters. Such parameters include the interface parameter $\epsilon,$ the number $k$ of eigenvalues for the Laplacian, the convexity splitting parameter $c,$ the (pseudo) time-step of the Allen--Cahn equation $\tau,$ as well as the 
correct stopping tolerance. One of the crucial questions is also the performance of the algorithms with respect to changes in the number of known or learned data. 
\subsection{Graphs}
Graph-based segmentation has been used for both UCI datasets \cite{Lichman:2013} and image based segmentation. We start with the scalar case for both a point set and an imaging problem. We later extend this to the multiclass segmentation. 
\subsubsection*{Scalar segmentation}
\begin{figure}
\centering
 \subfloat[Original image.]{\label{res:graph:scalar:im1_a}{
\includegraphics[width=.18\textwidth]{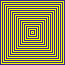}}
} \ 
\subfloat[Initial learned image.]{\label{res:graph:scalar:im1_b}{
\includegraphics[width=.18\textwidth]{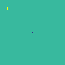}}
}\\
\subfloat[Solution of the smooth model.]{\label{res:graph:scalar:im1_c}{
\includegraphics[width=.18\textwidth]{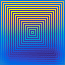}}
} \ 
\subfloat[Final segmentation using the smooth model.]{\label{res:graph:scalar:im1_d}{
\includegraphics[width=.18\textwidth]{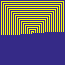}}
}\\
\subfloat[Solution of the non-smooth model.]{\label{res:graph:scalar:im1_e}{
\includegraphics[width=.18\textwidth]{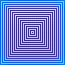}}
} \ 
\subfloat[Final segmentation using the non-smooth model.]{\label{res:graph:scalar:im1_f}{
\includegraphics[width=.18\textwidth]{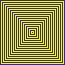}}
} 
\caption{Scalar image segmentation.}
\label{res:graph:scalar:im1}
\end{figure}

The first test we perform is based on the $65\times 65$ image given in Figure \ref{res:graph:scalar:im1}\subref{res:graph:scalar:im1_a}. 
This image consists of two colors - here given by dark blue and yellow. 
The learned information of the image used as initial state for the smooth and non-smooth model is shown in Figure \ref{res:graph:scalar:im1}\subref{res:graph:scalar:im1_b}. 
The known image information is given by one pixel in the dark blue part and three pixels in the yellow part. 
Hence, the known image information constitutes only of $0.0947\,\%$ of the whole image. 
The solution $u$ of the smooth model is presented in Figure \ref{res:graph:scalar:im1}\subref{res:graph:scalar:im1_c}, 
while Figure \ref{res:graph:scalar:im1}\subref{res:graph:scalar:im1_d} illustrates the final segmentation $\textup{sign}(u)$. 
The two corresponding results using the non-smooth model are given in Figure \ref{res:graph:scalar:im1}\subref{res:graph:scalar:im1_e} 
and \ref{res:graph:scalar:im1}\subref{res:graph:scalar:im1_f}. 
The chosen parameters are given as $\omega_{0} = 1$, $\eps = 0.5$, $\tau = 0.01$, $\nu=10^{-7}$, $c=2\epsilon^{-1}+\omega_{0}$, 
$R=21$, and $k=5$. Here, $R$ is the local scale for the graph Laplacian computation as used in (\ref{weightgraph_ZMP}). The computation of the eigenvalues is based on \texttt{svds} from MATLAB. As stopping criterion for the smooth and non-smooth model, we use
\begin{equation}
 \label{time_stop}
 \frac{\|u-{\bar u}\|}{\|{\bar u}\|}\leq\epsilon_{tol},
\end{equation}
where we set $\epsilon_{tol}=10^{-6}$, and we fix the maximum number of time steps to $t_{\max}=500$. 
Note that for the non-smooth model, we fix a sequence of penalty parameters $\{\nu_{q}\}_{q\in\mathbb{N}}$ with $\nu_{q}\to0$, and 
in each time step, we solve the problem $F(u_{\nu_{q}})=0$ for $q=1,\ldots,q_{\max}$ via a semi-smooth Newton method. 
In all examples, we use $\nu_{1}=10^{-1}\geq \nu_{2}=10^{-2}\geq\ldots\geq \nu_{q_{\max}}=10^{-7}$. 
Each semi-smooth Newton method is initialized by the approximate solution of the previous one. 
As stopping criterion for the semi-smooth Newton method, we use
\begin{equation}
 \label{SSN_stop}
\|F(u^{(l+1)})\|\leq\epsilon_{\textup{rel}}\|F(u^{(0)})\|+\epsilon_{\textup{abs}},\quad l=1,\ldots,l_{\max},
\end{equation}
where we set $l_{\max}=20,\,\epsilon_{\textup{rel}}=10^{-12}$, and $\epsilon_{\textup{abs}}=10^{-6}$. 
Finally, we solve the $k\times k$ systems of linear equations arising in each semi-smooth Newton step with MATLAB's backslash command.

The smooth model stops after $t_{\max}=500$ time steps with $\frac{\|u-{\bar u}\|}{\|{\bar u}\|}=9.5\cdot 10^{-4}$. 
The CPU time is $2.6s$ and the minimum and maximum value of the solution are $-1.165010$ and $1.296961$. 
The non-smooth model stops after $t_{\max}=500$ time steps with $\frac{\|u-{\bar u}\|}{\|{\bar u}\|}=8.6\cdot 10^{-3}$. 
The CPU time is $173.2s$ and the minimum and maximum value of the solution are $-1.000347$ and $1.000116$. 
We observe that the concentrations stay closer within the interval $[-1,1]$ when the non-smooth potential is used. 
Moreover, we clearly see from Figures \ref{res:graph:scalar:im1}\subref{res:graph:scalar:im1_c}--\ref{res:graph:scalar:im1}\subref{res:graph:scalar:im1_f} 
that the segmentation using the smooth model is either unsuccessful or has not finished after $500$ time steps. 
Before we will investigate this issue, we introduce the Figure of Certainty (FOC)
\begin{displaymath}
 \textup{FOC}(u_{\textup{orig}},p,q)=\frac{1}{n}\sum_{j=1}^{n}{\frac{1}{1+p\cdot|u_{\textup{orig}}(j)-u_{\textup{orig}}(s_{i})|^{q}}},
\end{displaymath}
a quality measurement proposed by Strasters and Gerbrand \cite{StrG91} in order to evaluate the quality of segmentation. 
Here, $u_{\textup{orig}}$ is the original image, $p,q>0$ are scaling parameters, $u_{\textup{orig}}(j)$ represents the intensity of pixel $j$, and 
$u_{\textup{orig}}(s_{i})$ is the intensity representing the segment that comprises pixel $j$.
It holds $\textup{FOC}\in(0,1]$, and the larger the FOC value is, the better is the segmentation. In the following, we use $p=q=0.5$ if not mentioned otherwise. 
In the smooth case (Figure \ref{res:graph:scalar:im1}\subref{res:graph:scalar:im1_d}), we obtain $\textup{FOC}=0.8186$ and in the non-smooth case (Figure \ref{res:graph:scalar:im1}\subref{res:graph:scalar:im1_f}) $\textup{FOC}=1$. 
Now, we come back to the above mentioned issue that the smooth model is either unsuccessful or has not finished after $500$ time steps. We repeat the same simulation for the smooth model with $t_{\max}=5000$ and $\epsilon_{tol}=10^{-14}$: The simulation stops after $1776$ time steps and a CPU time of $14.7s$ with $\frac{\|u-{\bar u}\|}{\|{\bar u}\|}=9.9\cdot 10^{-15}$ and $\textup{FOC}=0.8156$. 
Hence, the segmentation using the smooth model is unsuccessful with the used parameter set.\\

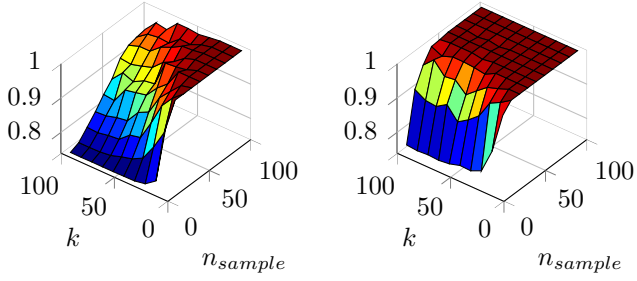
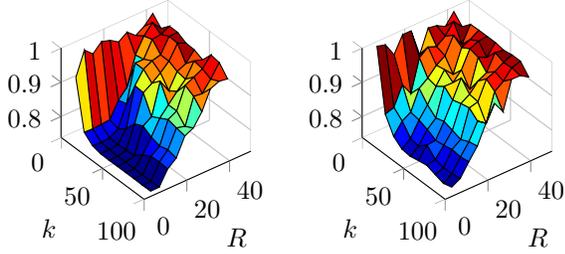
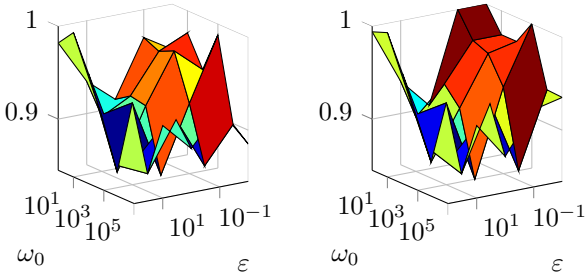
\begin{figure}
\centering
	\setlength\figureheight{0.3\linewidth} 
	\setlength\figurewidth{0.3\linewidth}
 \subfloat[Mean FOC values using the smooth model.]{\label{res:graph:scalar:im2_a}{
%
%
\begin{tikzpicture}

\begin{axis}[%
width=0.951\figurewidth,
height=\figureheight,
at={(0\figurewidth,0\figureheight)},
scale only axis,
point meta min=0.739090984508076,
point meta max=1,
every outer x axis line/.append style={black},
every x tick label/.append style={font=\color{black}},
xmin=0,
xmax=100,
tick align=outside,
xlabel={$n_{sample}$},
xmajorgrids,
every outer y axis line/.append style={black},
every y tick label/.append style={font=\color{black}},
ymin=0,
ymax=100,
ylabel={$k$},
ymajorgrids,
xlabel style={xshift=-0.4cm,yshift=0.1cm},
ylabel style={yshift=0.2cm,yshift=0.1cm},
every outer z axis line/.append style={black},
every z tick label/.append style={font=\color{black}},
zmin=0.739090984508076,
zmax=1,
zmajorgrids,
view={-52.1250163489018}{41.2526412730367},
axis background/.style={fill=white},
axis x line*=bottom,
axis y line*=left,
axis z line*=left
]

\addplot3[%
surf,
shader=flat corner,draw=black,z buffer=sort,colormap/jet,mesh/rows=10]
table[row sep=crcr, point meta=\thisrow{c}] {%
x	y	z	c\\
5	5	0.951313466630958	0.951313466630958\\
5	15	0.76862180911129	0.76862180911129\\
5	25	0.746357931574326	0.746357931574326\\
5	35	0.740466175991399	0.740466175991399\\
5	45	0.740007486498642	0.740007486498642\\
5	55	0.739502817182683	0.739502817182683\\
5	65	0.739418330555438	0.739418330555438\\
5	75	0.739695755899167	0.739695755899167\\
5	85	0.739090984508076	0.739090984508076\\
5	95	0.739488264579448	0.739488264579448\\
15	5	1	1\\
15	15	0.892031031420926	0.892031031420926\\
15	25	0.835541858585931	0.835541858585931\\
15	35	0.789117468469145	0.789117468469145\\
15	45	0.772979446523205	0.772979446523205\\
15	55	0.764731368645041	0.764731368645041\\
15	65	0.754699050465004	0.754699050465004\\
15	75	0.750861258456686	0.750861258456686\\
15	85	0.750002921748803	0.750002921748803\\
15	95	0.75016394408574	0.75016394408574\\
25	5	1	1\\
25	15	0.974301057081935	0.974301057081935\\
25	25	0.938406010750013	0.938406010750013\\
25	35	0.874240394566685	0.874240394566685\\
25	45	0.817306432718602	0.817306432718602\\
25	55	0.810135452034066	0.810135452034066\\
25	65	0.784128052638027	0.784128052638027\\
25	75	0.780964920533917	0.780964920533917\\
25	85	0.774869228221561	0.774869228221561\\
25	95	0.776691964675352	0.776691964675352\\
35	5	1	1\\
35	15	0.992299941030303	0.992299941030303\\
35	25	0.961425391042866	0.961425391042866\\
35	35	0.929352416228018	0.929352416228018\\
35	45	0.873675820751689	0.873675820751689\\
35	55	0.858029045178731	0.858029045178731\\
35	65	0.820208022924516	0.820208022924516\\
35	75	0.815025077837744	0.815025077837744\\
35	85	0.806207292574216	0.806207292574216\\
35	95	0.804095197944817	0.804095197944817\\
45	5	1	1\\
45	15	1	1\\
45	25	1	1\\
45	35	0.951671210408653	0.951671210408653\\
45	45	0.925548793588443	0.925548793588443\\
45	55	0.890843589063909	0.890843589063909\\
45	65	0.883958479084077	0.883958479084077\\
45	75	0.877385264030616	0.877385264030616\\
45	85	0.832697083547719	0.832697083547719\\
45	95	0.836441562044355	0.836441562044355\\
55	5	1	1\\
55	15	1	1\\
55	25	0.988502164973198	0.988502164973198\\
55	35	0.982962590013411	0.982962590013411\\
55	45	0.979942089557165	0.979942089557165\\
55	55	0.905488371116556	0.905488371116556\\
55	65	0.900563893013314	0.900563893013314\\
55	75	0.883517836569173	0.883517836569173\\
55	85	0.895633430438358	0.895633430438358\\
55	95	0.863254795668638	0.863254795668638\\
65	5	1	1\\
65	15	1	1\\
65	25	1	1\\
65	35	0.989309214294707	0.989309214294707\\
65	45	0.96591264016842	0.96591264016842\\
65	55	0.949982519030717	0.949982519030717\\
65	65	0.921360217639652	0.921360217639652\\
65	75	0.910886480269024	0.910886480269024\\
65	85	0.902208294576047	0.902208294576047\\
65	95	0.906060138368817	0.906060138368817\\
75	5	1	1\\
75	15	1	1\\
75	25	1	1\\
75	35	0.99506180712507	0.99506180712507\\
75	45	0.992916192155211	0.992916192155211\\
75	55	0.975923755559678	0.975923755559678\\
75	65	0.962366429562825	0.962366429562825\\
75	75	0.936714087345606	0.936714087345606\\
75	85	0.93668035685328	0.93668035685328\\
75	95	0.92561421141777	0.92561421141777\\
85	5	1	1\\
85	15	1	1\\
85	25	1	1\\
85	35	0.993063323186649	0.993063323186649\\
85	45	0.979498842375747	0.979498842375747\\
85	55	0.981611049426528	0.981611049426528\\
85	65	0.956374376989312	0.956374376989312\\
85	75	0.958825210591685	0.958825210591685\\
85	85	0.954114301409796	0.954114301409796\\
85	95	0.93181576418649	0.93181576418649\\
95	5	1	1\\
95	15	1	1\\
95	25	1	1\\
95	35	0.995906294415282	0.995906294415282\\
95	45	0.991234659217488	0.991234659217488\\
95	55	0.994731512880035	0.994731512880035\\
95	65	0.994943278804308	0.994943278804308\\
95	75	0.963244343026806	0.963244343026806\\
95	85	0.968657572879699	0.968657572879699\\
95	95	0.936020599023309	0.936020599023309\\
};
\end{axis}
\end{tikzpicture}%
}
} \ 
\subfloat[Mean FOC values using the non-smooth model.]{\label{res:graph:scalar:im2_b}{
%
%
\begin{tikzpicture}

\begin{axis}[%
width=0.951\figurewidth,
height=\figureheight,
at={(0\figurewidth,0\figureheight)},
scale only axis,
point meta min=0.739090984508076,
point meta max=1,
every outer x axis line/.append style={black},
every x tick label/.append style={font=\color{black}},
xmin=0,
xmax=100,
tick align=outside,
xlabel={$n_{sample}$},
xmajorgrids,
every outer y axis line/.append style={black},
every y tick label/.append style={font=\color{black}},
ymin=0,
ymax=100,
ylabel={$k$},
ymajorgrids,
xlabel style={xshift=-0.4cm,yshift=0.1cm},
ylabel style={yshift=0.2cm,yshift=0.1cm},
every outer z axis line/.append style={black},
every z tick label/.append style={font=\color{black}},
zmin=0.739090984508076,
zmax=1,
zmajorgrids,
view={-52.1250163489018}{41.2526412730367},
axis background/.style={fill=white},
axis x line*=bottom,
axis y line*=left,
axis z line*=left
]

\addplot3[%
surf,
shader=flat corner,draw=black,z buffer=sort,colormap/jet,mesh/rows=10]
table[row sep=crcr, point meta=\thisrow{c}] {%
x	y	z	c\\
5	5	0.973892033854432	0.973892033854432\\
5	15	0.855276562047452	0.855276562047452\\
5	25	0.784208397599669	0.784208397599669\\
5	35	0.77011979676768	0.77011979676768\\
5	45	0.771516512116297	0.771516512116297\\
5	55	0.762291407484932	0.762291407484932\\
5	65	0.758496473320345	0.758496473320345\\
5	75	0.760564075971558	0.760564075971558\\
5	85	0.758310734525097	0.758310734525097\\
5	95	0.76114013902732	0.76114013902732\\
15	5	1	1\\
15	15	0.979617043828648	0.979617043828648\\
15	25	0.963724648635395	0.963724648635395\\
15	35	0.912431428724907	0.912431428724907\\
15	45	0.886770715338283	0.886770715338283\\
15	55	0.866819709647738	0.866819709647738\\
15	65	0.903422375874079	0.903422375874079\\
15	75	0.883820306322383	0.883820306322383\\
15	85	0.887115825614086	0.887115825614086\\
15	95	0.905029939193036	0.905029939193036\\
25	5	1	1\\
25	15	1	1\\
25	25	0.987818275590459	0.987818275590459\\
25	35	0.977561298065418	0.977561298065418\\
25	45	0.953279635876489	0.953279635876489\\
25	55	0.927952551393326	0.927952551393326\\
25	65	0.957501684161492	0.957501684161492\\
25	75	0.953061090666701	0.953061090666701\\
25	85	0.987629879427188	0.987629879427188\\
25	95	0.958348476030697	0.958348476030697\\
35	5	1	1\\
35	15	1	1\\
35	25	1	1\\
35	35	1	1\\
35	45	0.98393609945231	0.98393609945231\\
35	55	0.989208871183681	0.989208871183681\\
35	65	0.9917156678185	0.9917156678185\\
35	75	0.99179919176769	0.99179919176769\\
35	85	0.980047145299823	0.980047145299823\\
35	95	0.994137516663546	0.994137516663546\\
45	5	1	1\\
45	15	1	1\\
45	25	1	1\\
45	35	1	1\\
45	45	1	1\\
45	55	0.994785022154792	0.994785022154792\\
45	65	1	1\\
45	75	1	1\\
45	85	1	1\\
45	95	0.996224570334465	0.996224570334465\\
55	5	1	1\\
55	15	1	1\\
55	25	1	1\\
55	35	1	1\\
55	45	1	1\\
55	55	1	1\\
55	65	1	1\\
55	75	1	1\\
55	85	1	1\\
55	95	1	1\\
65	5	1	1\\
65	15	1	1\\
65	25	1	1\\
65	35	1	1\\
65	45	1	1\\
65	55	0.994785022154793	0.994785022154793\\
65	65	1	1\\
65	75	0.99515564406481	0.99515564406481\\
65	85	1	1\\
65	95	1	1\\
75	5	1	1\\
75	15	1	1\\
75	25	1	1\\
75	35	1	1\\
75	45	1	1\\
75	55	1	1\\
75	65	1	1\\
75	75	1	1\\
75	85	1	1\\
75	95	1	1\\
85	5	1	1\\
85	15	1	1\\
85	25	1	1\\
85	35	1	1\\
85	45	1	1\\
85	55	1	1\\
85	65	1	1\\
85	75	1	1\\
85	85	1	1\\
85	95	1	1\\
95	5	1	1\\
95	15	1	1\\
95	25	1	1\\
95	35	1	1\\
95	45	1	1\\
95	55	1	1\\
95	65	1	1\\
95	75	1	1\\
95	85	1	1\\
95	95	1	1\\
};
\end{axis}
\end{tikzpicture}%
}
}\\
\subfloat[Mean FOC values using the smooth model.]{\label{res:graph:scalar:im2_c}{
%
%
\begin{tikzpicture}

\begin{axis}[%
width=0.951\figurewidth,
height=\figureheight,
at={(0\figurewidth,0\figureheight)},
scale only axis,
point meta min=0.739100583062163,
point meta max=1,
every outer x axis line/.append style={black},
every x tick label/.append style={font=\color{black}},
xmin=0,
xmax=100,
tick align=outside,
xlabel={$k$},
xmajorgrids,
every outer y axis line/.append style={black},
every y tick label/.append style={font=\color{black}},
ymin=0,
ymax=50,
ylabel={$R$},
ymajorgrids,
xlabel style={yshift=0.3cm},
ylabel style={yshift=0.2cm},
every outer z axis line/.append style={black},
every z tick label/.append style={font=\color{black}},
zmin=0.739100583062163,
zmax=1,
zmajorgrids,
view={52.1250163489018}{41.2526412730367},
axis background/.style={fill=white},
axis x line*=bottom,
axis y line*=left,
axis z line*=left
]

\addplot3[%
surf,
shader=flat corner,draw=black,z buffer=sort,colormap/jet,mesh/rows=10]
table[row sep=crcr, point meta=\thisrow{c}] {%
x	y	z	c\\
5	5	0.908647028324523	0.908647028324523\\
5	9	0.977441308294226	0.977441308294226\\
5	13	0.955307419635964	0.955307419635964\\
5	17	0.979627013884469	0.979627013884469\\
5	21	0.972345769894601	0.972345769894601\\
5	25	0.956308876608047	0.956308876608047\\
5	29	0.973747682997731	0.973747682997731\\
5	33	0.952644735435644	0.952644735435644\\
5	37	0.9683801557568	0.9683801557568\\
5	41	0.994475426570816	0.994475426570816\\
15	5	0.767281362063869	0.767281362063869\\
15	9	0.766183262453509	0.766183262453509\\
15	13	0.767884333792258	0.767884333792258\\
15	17	0.770490153095845	0.770490153095845\\
15	21	0.815741484209374	0.815741484209374\\
15	25	0.84285924960969	0.84285924960969\\
15	29	0.925378971777747	0.925378971777747\\
15	33	0.969245629120098	0.969245629120098\\
15	37	0.996554656799176	0.996554656799176\\
15	41	0.96896423870272	0.96896423870272\\
25	5	0.746029180218516	0.746029180218516\\
25	9	0.744320781046055	0.744320781046055\\
25	13	0.746846728290216	0.746846728290216\\
25	17	0.761360457205291	0.761360457205291\\
25	21	0.823560180438305	0.823560180438305\\
25	25	0.930896942255566	0.930896942255566\\
25	29	0.930550898520935	0.930550898520935\\
25	33	0.977348125029996	0.977348125029996\\
25	37	0.980466020059747	0.980466020059747\\
25	41	0.993552588007642	0.993552588007642\\
35	5	0.740025469694225	0.740025469694225\\
35	9	0.741717059766941	0.741717059766941\\
35	13	0.745179309114837	0.745179309114837\\
35	17	0.780646640999114	0.780646640999114\\
35	21	0.894551491630327	0.894551491630327\\
35	25	0.930811305808726	0.930811305808726\\
35	29	0.944136869049682	0.944136869049682\\
35	33	0.950527445569576	0.950527445569576\\
35	37	0.992836938512328	0.992836938512328\\
35	41	0.993380705763777	0.993380705763777\\
45	5	0.739496239187771	0.739496239187771\\
45	9	0.739789194899015	0.739789194899015\\
45	13	0.74640026295672	0.74640026295672\\
45	17	0.812396169686645	0.812396169686645\\
45	21	0.892204153058812	0.892204153058812\\
45	25	0.908190478935679	0.908190478935679\\
45	29	0.927250879977925	0.927250879977925\\
45	33	0.950453420726619	0.950453420726619\\
45	37	0.991082965214713	0.991082965214713\\
45	41	0.980319282936227	0.980319282936227\\
55	5	0.739677154939405	0.739677154939405\\
55	9	0.740266032612206	0.740266032612206\\
55	13	0.749657947901113	0.749657947901113\\
55	17	0.824234221646169	0.824234221646169\\
55	21	0.84039135705234	0.84039135705234\\
55	25	0.932246768178196	0.932246768178196\\
55	29	0.893405504397344	0.893405504397344\\
55	33	0.938051146272026	0.938051146272026\\
55	37	0.949582838290155	0.949582838290155\\
55	41	0.978792211989094	0.978792211989094\\
65	5	0.739838971720442	0.739838971720442\\
65	9	0.739677994309889	0.739677994309889\\
65	13	0.769427528063703	0.769427528063703\\
65	17	0.809937128583943	0.809937128583943\\
65	21	0.853996942523018	0.853996942523018\\
65	25	0.881346889867136	0.881346889867136\\
65	29	0.908813896623243	0.908813896623243\\
65	33	0.933496800544789	0.933496800544789\\
65	37	0.975273790596561	0.975273790596561\\
65	41	0.967518274378396	0.967518274378396\\
75	5	0.739988001966642	0.739988001966642\\
75	9	0.7395088240893	0.7395088240893\\
75	13	0.782519781143516	0.782519781143516\\
75	17	0.802079755137725	0.802079755137725\\
75	21	0.867818413532262	0.867818413532262\\
75	25	0.881609852592432	0.881609852592432\\
75	29	0.945157335691326	0.945157335691326\\
75	33	0.957850056297195	0.957850056297195\\
75	37	0.970626430793999	0.970626430793999\\
75	41	1	1\\
85	5	0.739225928976268	0.739225928976268\\
85	9	0.739100583062163	0.739100583062163\\
85	13	0.784554163155349	0.784554163155349\\
85	17	0.809035415550599	0.809035415550599\\
85	21	0.861310457982943	0.861310457982943\\
85	25	0.85479713509197	0.85479713509197\\
85	29	0.931302763845219	0.931302763845219\\
85	33	0.927779713596513	0.927779713596513\\
85	37	0.958457245877756	0.958457245877756\\
85	41	0.995897646937777	0.995897646937777\\
95	5	0.739756097448497	0.739756097448497\\
95	9	0.740336438580116	0.740336438580116\\
95	13	0.78233299907456	0.78233299907456\\
95	17	0.802833390101814	0.802833390101814\\
95	21	0.855948341840978	0.855948341840978\\
95	25	0.86830328840758	0.86830328840758\\
95	29	0.904272620754924	0.904272620754924\\
95	33	0.95980336230782	0.95980336230782\\
95	37	0.970268786710674	0.970268786710674\\
95	41	0.966348563205473	0.966348563205473\\
};
\end{axis}
\end{tikzpicture}%
}
} \ 
\subfloat[Mean FOC values using the non-smooth model.]{\label{res:graph:scalar:im2_d}{
%
%
\begin{tikzpicture}

\begin{axis}[%
width=0.951\figurewidth,
height=\figureheight,
at={(0\figurewidth,0\figureheight)},
scale only axis,
point meta min=0.739100583062163,
point meta max=1,
every outer x axis line/.append style={black},
every x tick label/.append style={font=\color{black}},
xmin=0,
xmax=100,
tick align=outside,
xlabel={$k$},
xmajorgrids,
every outer y axis line/.append style={black},
every y tick label/.append style={font=\color{black}},
ymin=0,
ymax=50,
ylabel={$R$},
ymajorgrids,
xlabel style={yshift=0.3cm},
ylabel style={yshift=0.2cm},
every outer z axis line/.append style={black},
every z tick label/.append style={font=\color{black}},
zmin=0.739100583062163,
zmax=1,
zmajorgrids,
view={52.1250163489018}{41.2526412730367},
axis background/.style={fill=white},
axis x line*=bottom,
axis y line*=left,
axis z line*=left
]

\addplot3[%
surf,
shader=flat corner,draw=black,z buffer=sort,colormap/jet,mesh/rows=10]
table[row sep=crcr, point meta=\thisrow{c}] {%
x	y	z	c\\
5	5	1	1\\
5	9	0.992757192566323	0.992757192566323\\
5	13	0.95299418710358	0.95299418710358\\
5	17	1	1\\
5	21	1	1\\
5	25	0.919827186630309	0.919827186630309\\
5	29	0.985824002300168	0.985824002300168\\
5	33	0.981561275732246	0.981561275732246\\
5	37	0.967665332274955	0.967665332274955\\
5	41	1	1\\
15	5	0.820382283856192	0.820382283856192\\
15	9	0.828461171582264	0.828461171582264\\
15	13	0.883683278919528	0.883683278919528\\
15	17	0.847015615702209	0.847015615702209\\
15	21	0.939996718100633	0.939996718100633\\
15	25	0.959733079487385	0.959733079487385\\
15	29	0.949898300676964	0.949898300676964\\
15	33	0.968520276632538	0.968520276632538\\
15	37	1	1\\
15	41	0.973892033854432	0.973892033854432\\
25	5	0.773465419614183	0.773465419614183\\
25	9	0.80672803707959	0.80672803707959\\
25	13	0.821797501755406	0.821797501755406\\
25	17	0.880118565207143	0.880118565207143\\
25	21	0.898797964370085	0.898797964370085\\
25	25	0.924413056069601	0.924413056069601\\
25	29	0.940692413370287	0.940692413370287\\
25	33	0.989415053103326	0.989415053103326\\
25	37	0.951145865155065	0.951145865155065\\
25	41	0.99888826499939	0.99888826499939\\
35	5	0.754244221492225	0.754244221492225\\
35	9	0.774737737856302	0.774737737856302\\
35	13	0.817111256787008	0.817111256787008\\
35	17	0.850163647669986	0.850163647669986\\
35	21	0.892635785006214	0.892635785006214\\
35	25	0.929522104599861	0.929522104599861\\
35	29	0.90472907296632	0.90472907296632\\
35	33	0.971923405685706	0.971923405685706\\
35	37	0.983592111009838	0.983592111009838\\
35	41	1	1\\
45	5	0.751194792209025	0.751194792209025\\
45	9	0.770001098205446	0.770001098205446\\
45	13	0.807146853979054	0.807146853979054\\
45	17	0.829622439428017	0.829622439428017\\
45	21	0.910413984985808	0.910413984985808\\
45	25	0.909655622570839	0.909655622570839\\
45	29	0.935965580312631	0.935965580312631\\
45	33	0.986331281652801	0.986331281652801\\
45	37	0.997068162385367	0.997068162385367\\
45	41	1	1\\
55	5	0.756508608736196	0.756508608736196\\
55	9	0.790658903853711	0.790658903853711\\
55	13	0.793009225839471	0.793009225839471\\
55	17	0.837589618056912	0.837589618056912\\
55	21	0.88858886652771	0.88858886652771\\
55	25	0.941224033356275	0.941224033356275\\
55	29	0.896863359923291	0.896863359923291\\
55	33	0.979279003138858	0.979279003138858\\
55	37	0.967547653472023	0.967547653472023\\
55	41	1	1\\
65	5	0.760919609308093	0.760919609308093\\
65	9	0.766909696689783	0.766909696689783\\
65	13	0.796969488080979	0.796969488080979\\
65	17	0.829629522380498	0.829629522380498\\
65	21	0.896498748568635	0.896498748568635\\
65	25	0.911603983992049	0.911603983992049\\
65	29	0.93924883266114	0.93924883266114\\
65	33	0.952552060008245	0.952552060008245\\
65	37	1	1\\
65	41	0.995653900999384	0.995653900999384\\
75	5	0.752249130116025	0.752249130116025\\
75	9	0.76109156073007	0.76109156073007\\
75	13	0.803072827987199	0.803072827987199\\
75	17	0.820798094015757	0.820798094015757\\
75	21	0.901933417159223	0.901933417159223\\
75	25	0.908489178900357	0.908489178900357\\
75	29	0.963674685012822	0.963674685012822\\
75	33	0.973461882028436	0.973461882028436\\
75	37	0.989383591829055	0.989383591829055\\
75	41	1	1\\
85	5	0.744487599606582	0.744487599606582\\
85	9	0.761217662689801	0.761217662689801\\
85	13	0.799299060781611	0.799299060781611\\
85	17	0.832639948596173	0.832639948596173\\
85	21	0.881267094522495	0.881267094522495\\
85	25	0.863161670925596	0.863161670925596\\
85	29	0.945424576868262	0.945424576868262\\
85	33	0.963822399716907	0.963822399716907\\
85	37	0.991436816934002	0.991436816934002\\
85	41	1	1\\
95	5	0.751542948541672	0.751542948541672\\
95	9	0.765800321254654	0.765800321254654\\
95	13	0.80342880287985	0.80342880287985\\
95	17	0.837684222921829	0.837684222921829\\
95	21	0.893196339988514	0.893196339988514\\
95	25	0.93976020176709	0.93976020176709\\
95	29	0.904291302099251	0.904291302099251\\
95	33	0.987576102334731	0.987576102334731\\
95	37	0.98673818227362	0.98673818227362\\
95	41	0.971791826288928	0.971791826288928\\
};
\end{axis}
\end{tikzpicture}%
}
}\\
\subfloat[Mean FOC values using the smooth model.]{\label{res:graph:scalar:im2_e}{
%
%
\begin{tikzpicture}

\begin{axis}[%
width=0.951\figurewidth,
height=\figureheight,
at={(0\figurewidth,0\figureheight)},
scale only axis,
every outer x axis line/.append style={black},
every x tick label/.append style={font=\color{black}},
xmode=log,
xmin=0.01,
xmax=100,
xminorticks=true,
tick align=outside,
xlabel={$\varepsilon$},
xmajorgrids,
xminorgrids,
every outer y axis line/.append style={black},
every y tick label/.append style={font=\color{black}},
ymode=log,
ymin=1,
ymax=100000,
yminorticks=true,
ylabel={$\omega_{0}$},
ymajorgrids,
yminorgrids,
xlabel style={yshift=0.1cm},
ylabel style={yshift=0.1cm},
every outer z axis line/.append style={black},
every z tick label/.append style={font=\color{black}},
zmin=0.844996245842338,
zmax=1,
zmajorgrids,
view={146.30993247402}{15.501359566937},
axis background/.style={fill=white},
axis x line*=bottom,
axis y line*=left,
axis z line*=left
]

\addplot3[%
surf,
shader=flat corner,draw=black,z buffer=sort,colormap/jet,mesh/rows=5]
table[row sep=crcr, point meta=\thisrow{c}] {%
x	y	z	c\\
0.01	1	0.951083743293137	0.951083743293137\\
0.01	10	0.97592876055978	0.97592876055978\\
0.01	100	0.943765518998404	0.943765518998404\\
0.01	1000	0.988062525184644	0.988062525184644\\
0.01	10000	0.899016574465338	0.899016574465338\\
0.01	100000	0.885099874674417	0.885099874674417\\
0.1	1	0.969728947256179	0.969728947256179\\
0.1	10	0.962113407951726	0.962113407951726\\
0.1	100	0.969767914868219	0.969767914868219\\
0.1	1000	0.920590742989557	0.920590742989557\\
0.1	10000	0.860423985648125	0.860423985648125\\
0.1	100000	0.927960808212649	0.927960808212649\\
1	1	0.909371915500998	0.909371915500998\\
1	10	0.920111303822371	0.920111303822371\\
1	100	0.916812812217407	0.916812812217407\\
1	1000	0.849243546642858	0.849243546642858\\
1	10000	0.916040722212022	0.916040722212022\\
1	100000	0.892268314348899	0.892268314348899\\
10	1	0.934567719215565	0.934567719215565\\
10	10	0.903938943290838	0.903938943290838\\
10	100	0.851366072211552	0.851366072211552\\
10	1000	0.932861795390283	0.932861795390283\\
10	10000	0.864075003084814	0.864075003084814\\
10	100000	0.922798590393238	0.922798590393238\\
100	1	0.981746961490086	0.981746961490086\\
100	10	1	1\\
100	100	0.962888041999853	0.962888041999853\\
100	1000	0.927025682547893	0.927025682547893\\
100	10000	0.878277733946626	0.878277733946626\\
100	100000	0.889944278017279	0.889944278017279\\
};
\end{axis}
\end{tikzpicture}%
}
} \ 
\subfloat[Mean FOC values using the non-smooth model.]{\label{res:graph:scalar:im2_f}{
%
%
\begin{tikzpicture}

\begin{axis}[%
width=0.951\figurewidth,
height=\figureheight,
at={(0\figurewidth,0\figureheight)},
scale only axis,
every outer x axis line/.append style={black},
every x tick label/.append style={font=\color{black}},
xmode=log,
xmin=0.01,
xmax=100,
xminorticks=true,
tick align=outside,
xlabel={$\varepsilon$},
xmajorgrids,
xminorgrids,
every outer y axis line/.append style={black},
every y tick label/.append style={font=\color{black}},
ymode=log,
ymin=1,
ymax=100000,
yminorticks=true,
ylabel={$\omega_{0}$},
ymajorgrids,
yminorgrids,
xlabel style={yshift=0.1cm},
ylabel style={yshift=0.1cm},
every outer z axis line/.append style={black},
every z tick label/.append style={font=\color{black}},
zmin=0.844996245842338,
zmax=1,
zmajorgrids,
view={146.30993247402}{15.501359566937},
axis background/.style={fill=white},
axis x line*=bottom,
axis y line*=left,
axis z line*=left
]

\addplot3[%
surf,
shader=flat corner,draw=black,z buffer=sort,colormap/jet,mesh/rows=5]
table[row sep=crcr, point meta=\thisrow{c}] {%
x	y	z	c\\
0.01	1	1	1\\
0.01	10	0.973892033854432	0.973892033854432\\
0.01	100	0.973892033854432	0.973892033854432\\
0.01	1000	1	1\\
0.01	10000	0.936155830911752	0.936155830911752\\
0.01	100000	0.935164825021424	0.935164825021424\\
0.1	1	1	1\\
0.1	10	0.973892033854432	0.973892033854432\\
0.1	100	0.96704137716603	0.96704137716603\\
0.1	1000	0.93806587391534	0.93806587391534\\
0.1	10000	0.855312333440178	0.855312333440178\\
0.1	100000	0.927910164970796	0.927910164970796\\
1	1	0.902870861657948	0.902870861657948\\
1	10	0.93291921880972	0.93291921880972\\
1	100	0.914233579243226	0.914233579243226\\
1	1000	0.844996245842338	0.844996245842338\\
1	10000	0.93322716949464	0.93322716949464\\
1	100000	0.883716653506793	0.883716653506793\\
10	1	0.934567719215565	0.934567719215565\\
10	10	0.903460026318706	0.903460026318706\\
10	100	0.862111439689316	0.862111439689316\\
10	1000	0.932861795390283	0.932861795390283\\
10	10000	0.863980703454197	0.863980703454197\\
10	100000	0.922798590393238	0.922798590393238\\
100	1	0.993732988011866	0.993732988011866\\
100	10	1	1\\
100	100	0.962888041999853	0.962888041999853\\
100	1000	0.927006338799586	0.927006338799586\\
100	10000	0.893416098878644	0.893416098878644\\
100	100000	0.889944278017279	0.889944278017279\\
};
\end{axis}
\end{tikzpicture}%
}
}
\caption{Comparison of the smooth and non-smooth model: The mean of the FOC values for the smooth (left column) and non-smooth (right column) model for varying parameters. For each $(x,y)$ pair, we have taken $10$ runs with randomly chosen samples.}
\label{res:graph:scalar:im2}
\end{figure}

\begin{figure}
\centering
	\setlength\figureheight{0.3\linewidth} 
	\setlength\figurewidth{0.3\linewidth}
\subfloat[Difference of the results in Figure \protect\ref{res:graph:scalar:im2}\protect\subref{res:graph:scalar:im2_a} and \protect\ref{res:graph:scalar:im2}\protect\subref{res:graph:scalar:im2_b}.]{\label{res:graph:scalar:im3_a}{
%
%
\begin{tikzpicture}

\begin{axis}[%
width=0.851\figurewidth,
height=\figureheight,
at={(0\figurewidth,0\figureheight)},
scale only axis,
every outer x axis line/.append style={black},
every x tick label/.append style={font=\color{black}},
xmin=0,
xmax=100,
tick align=outside,
xlabel={$n_{sample}$},
xmajorgrids,
every outer y axis line/.append style={black},
every y tick label/.append style={font=\color{black}},
ymin=0,
ymax=100,
ylabel={$k$},
ymajorgrids,
xlabel style={xshift=-0.4cm,yshift=0.1cm},
ylabel style={yshift=0.2cm,yshift=0.1cm},
every outer z axis line/.append style={black},
every z tick label/.append style={font=\color{black}},
zmin=-0.25,
zmax=0,
zmajorgrids,
view={-52.1250163489018}{41.2526412730367},
axis background/.style={fill=white},
axis x line*=bottom,
axis y line*=left,
axis z line*=left
]

\addplot3[%
surf,
shader=flat corner,draw=black,z buffer=sort,colormap/jet,mesh/rows=10]
table[row sep=crcr, point meta=\thisrow{c}] {%
x	y	z	c\\
5	5	-0.0225785672234741	-0.0225785672234741\\
5	15	-0.086654752936162	-0.086654752936162\\
5	25	-0.037850466025343	-0.037850466025343\\
5	35	-0.029653620776281	-0.029653620776281\\
5	45	-0.0315090256176558	-0.0315090256176558\\
5	55	-0.0227885903022489	-0.0227885903022489\\
5	65	-0.0190781427649069	-0.0190781427649069\\
5	75	-0.0208683200723908	-0.0208683200723908\\
5	85	-0.0192197500170213	-0.0192197500170213\\
5	95	-0.0216518744478721	-0.0216518744478721\\
15	5	0	0\\
15	15	-0.0875860124077219	-0.0875860124077219\\
15	25	-0.128182790049464	-0.128182790049464\\
15	35	-0.123313960255763	-0.123313960255763\\
15	45	-0.113791268815078	-0.113791268815078\\
15	55	-0.102088341002697	-0.102088341002697\\
15	65	-0.148723325409074	-0.148723325409074\\
15	75	-0.132959047865697	-0.132959047865697\\
15	85	-0.137112903865283	-0.137112903865283\\
15	95	-0.154865995107296	-0.154865995107296\\
25	5	0	0\\
25	15	-0.0256989429180651	-0.0256989429180651\\
25	25	-0.0494122648404458	-0.0494122648404458\\
25	35	-0.103320903498733	-0.103320903498733\\
25	45	-0.135973203157887	-0.135973203157887\\
25	55	-0.11781709935926	-0.11781709935926\\
25	65	-0.173373631523465	-0.173373631523465\\
25	75	-0.172096170132783	-0.172096170132783\\
25	85	-0.212760651205627	-0.212760651205627\\
25	95	-0.181656511355345	-0.181656511355345\\
35	5	0	0\\
35	15	-0.00770005896969705	-0.00770005896969705\\
35	25	-0.0385746089571345	-0.0385746089571345\\
35	35	-0.0706475837719818	-0.0706475837719818\\
35	45	-0.110260278700621	-0.110260278700621\\
35	55	-0.13117982600495	-0.13117982600495\\
35	65	-0.171507644893984	-0.171507644893984\\
35	75	-0.176774113929946	-0.176774113929946\\
35	85	-0.173839852725607	-0.173839852725607\\
35	95	-0.190042318718729	-0.190042318718729\\
45	5	0	0\\
45	15	0	0\\
45	25	0	0\\
45	35	-0.048328789591347	-0.048328789591347\\
45	45	-0.0744512064115572	-0.0744512064115572\\
45	55	-0.103941433090883	-0.103941433090883\\
45	65	-0.116041520915923	-0.116041520915923\\
45	75	-0.122614735969384	-0.122614735969384\\
45	85	-0.167302916452281	-0.167302916452281\\
45	95	-0.15978300829011	-0.15978300829011\\
55	5	0	0\\
55	15	0	0\\
55	25	-0.0114978350268018	-0.0114978350268018\\
55	35	-0.0170374099865889	-0.0170374099865889\\
55	45	-0.020057910442835	-0.020057910442835\\
55	55	-0.0945116288834443	-0.0945116288834443\\
55	65	-0.0994361069866864	-0.0994361069866864\\
55	75	-0.116482163430827	-0.116482163430827\\
55	85	-0.104366569561642	-0.104366569561642\\
55	95	-0.136745204331362	-0.136745204331362\\
65	5	0	0\\
65	15	0	0\\
65	25	0	0\\
65	35	-0.0106907857052931	-0.0106907857052931\\
65	45	-0.0340873598315804	-0.0340873598315804\\
65	55	-0.044802503124076	-0.044802503124076\\
65	65	-0.078639782360348	-0.078639782360348\\
65	75	-0.0842691637957861	-0.0842691637957861\\
65	85	-0.097791705423953	-0.097791705423953\\
65	95	-0.0939398616311828	-0.0939398616311828\\
75	5	0	0\\
75	15	0	0\\
75	25	0	0\\
75	35	-0.00493819287493003	-0.00493819287493003\\
75	45	-0.00708380784478879	-0.00708380784478879\\
75	55	-0.024076244440322	-0.024076244440322\\
75	65	-0.0376335704371751	-0.0376335704371751\\
75	75	-0.0632859126543937	-0.0632859126543937\\
75	85	-0.0633196431467196	-0.0633196431467196\\
75	95	-0.0743857885822296	-0.0743857885822296\\
85	5	0	0\\
85	15	0	0\\
85	25	0	0\\
85	35	-0.00693667681335097	-0.00693667681335097\\
85	45	-0.0205011576242526	-0.0205011576242526\\
85	55	-0.0183889505734719	-0.0183889505734719\\
85	65	-0.043625623010688	-0.043625623010688\\
85	75	-0.0411747894083145	-0.0411747894083145\\
85	85	-0.0458856985902036	-0.0458856985902036\\
85	95	-0.0681842358135099	-0.0681842358135099\\
95	5	0	0\\
95	15	0	0\\
95	25	0	0\\
95	35	-0.00409370558471844	-0.00409370558471844\\
95	45	-0.00876534078251168	-0.00876534078251168\\
95	55	-0.00526848711996464	-0.00526848711996464\\
95	65	-0.00505672119569189	-0.00505672119569189\\
95	75	-0.0367556569731942	-0.0367556569731942\\
95	85	-0.0313424271203011	-0.0313424271203011\\
95	95	-0.0639794009766905	-0.0639794009766905\\
};
\end{axis}
\end{tikzpicture}%
}
} \ 
\subfloat[Difference of the results in Figure \protect\ref{res:graph:scalar:im2}\protect\subref{res:graph:scalar:im2_c} and \protect\ref{res:graph:scalar:im2}\protect\subref{res:graph:scalar:im2_d}.]{\label{res:graph:scalar:im3_b}{
%
%
\begin{tikzpicture}

\begin{axis}[%
width=0.851\figurewidth,
height=\figureheight,
at={(0\figurewidth,0\figureheight)},
scale only axis,
every outer x axis line/.append style={black},
every x tick label/.append style={font=\color{black}},
xmin=0,
xmax=100,
tick align=outside,
xlabel={$k$},
xmajorgrids,
every outer y axis line/.append style={black},
every y tick label/.append style={font=\color{black}},
ymin=0,
ymax=50,
ylabel={$R$},
ymajorgrids,
xlabel style={yshift=0.3cm},
ylabel style={yshift=0.2cm},
every outer z axis line/.append style={black},
every z tick label/.append style={font=\color{black}},
zmin=-0.15,
zmax=0.05,
zmajorgrids,
view={52.1250163489018}{41.2526412730367},
axis background/.style={fill=white},
axis x line*=bottom,
axis y line*=left,
axis z line*=left
]

\addplot3[%
surf,
shader=flat corner,draw=black,z buffer=sort,colormap/jet,mesh/rows=10]
table[row sep=crcr, point meta=\thisrow{c}] {%
x	y	z	c\\
5	5	-0.0913529716754772	-0.0913529716754772\\
5	9	-0.0153158842720973	-0.0153158842720973\\
5	13	0.00231323253238358	0.00231323253238358\\
5	17	-0.0203729861155312	-0.0203729861155312\\
5	21	-0.0276542301053995	-0.0276542301053995\\
5	25	0.0364816899777376	0.0364816899777376\\
5	29	-0.0120763193024368	-0.0120763193024368\\
5	33	-0.0289165402966015	-0.0289165402966015\\
5	37	0.000714823481845595	0.000714823481845595\\
5	41	-0.00552457342918444	-0.00552457342918444\\
15	5	-0.0531009217923228	-0.0531009217923228\\
15	9	-0.0622779091287551	-0.0622779091287551\\
15	13	-0.115798945127269	-0.115798945127269\\
15	17	-0.0765254626063647	-0.0765254626063647\\
15	21	-0.124255233891259	-0.124255233891259\\
15	25	-0.116873829877695	-0.116873829877695\\
15	29	-0.0245193288992167	-0.0245193288992167\\
15	33	0.000725352487559894	0.000725352487559894\\
15	37	-0.00344534320082401	-0.00344534320082401\\
15	41	-0.00492779515171216	-0.00492779515171216\\
25	5	-0.0274362393956672	-0.0274362393956672\\
25	9	-0.0624072560335346	-0.0624072560335346\\
25	13	-0.0749507734651897	-0.0749507734651897\\
25	17	-0.118758108001852	-0.118758108001852\\
25	21	-0.0752377839317802	-0.0752377839317802\\
25	25	0.00648388618596552	0.00648388618596552\\
25	29	-0.010141514849352	-0.010141514849352\\
25	33	-0.0120669280733305	-0.0120669280733305\\
25	37	0.0293201549046826	0.0293201549046826\\
25	41	-0.00533567699174786	-0.00533567699174786\\
35	5	-0.0142187517980003	-0.0142187517980003\\
35	9	-0.0330206780893613	-0.0330206780893613\\
35	13	-0.0719319476721709	-0.0719319476721709\\
35	17	-0.0695170066708718	-0.0695170066708718\\
35	21	0.00191570662411245	0.00191570662411245\\
35	25	0.00128920120886444	0.00128920120886444\\
35	29	0.0394077960833614	0.0394077960833614\\
35	33	-0.0213959601161298	-0.0213959601161298\\
35	37	0.00924482750248934	0.00924482750248934\\
35	41	-0.00661929423622287	-0.00661929423622287\\
45	5	-0.0116985530212541	-0.0116985530212541\\
45	9	-0.0302119033064315	-0.0302119033064315\\
45	13	-0.0607465910223335	-0.0607465910223335\\
45	17	-0.0172262697413723	-0.0172262697413723\\
45	21	-0.0182098319269963	-0.0182098319269963\\
45	25	-0.00146514363516026	-0.00146514363516026\\
45	29	-0.00871470033470545	-0.00871470033470545\\
45	33	-0.0358778609261821	-0.0358778609261821\\
45	37	-0.00598519717065371	-0.00598519717065371\\
45	41	-0.0196807170637726	-0.0196807170637726\\
55	5	-0.0168314537967913	-0.0168314537967913\\
55	9	-0.0503928712415049	-0.0503928712415049\\
55	13	-0.0433512779383584	-0.0433512779383584\\
55	17	-0.0133553964107423	-0.0133553964107423\\
55	21	-0.0481975094753705	-0.0481975094753705\\
55	25	-0.00897726517807995	-0.00897726517807995\\
55	29	-0.00345785552594724	-0.00345785552594724\\
55	33	-0.0412278568668317	-0.0412278568668317\\
55	37	-0.0179648151818682	-0.0179648151818682\\
55	41	-0.0212077880109064	-0.0212077880109064\\
65	5	-0.0210806375876501	-0.0210806375876501\\
65	9	-0.0272317023798935	-0.0272317023798935\\
65	13	-0.0275419600172754	-0.0275419600172754\\
65	17	-0.0196923937965556	-0.0196923937965556\\
65	21	-0.042501806045617	-0.042501806045617\\
65	25	-0.030257094124913	-0.030257094124913\\
65	29	-0.0304349360378963	-0.0304349360378963\\
65	33	-0.0190552594634555	-0.0190552594634555\\
65	37	-0.0247262094034391	-0.0247262094034391\\
65	41	-0.0281356266209883	-0.0281356266209883\\
75	5	-0.0122611281493833	-0.0122611281493833\\
75	9	-0.02158273664077	-0.02158273664077\\
75	13	-0.0205530468436831	-0.0205530468436831\\
75	17	-0.0187183388780315	-0.0187183388780315\\
75	21	-0.0341150036269608	-0.0341150036269608\\
75	25	-0.0268793263079248	-0.0268793263079248\\
75	29	-0.0185173493214958	-0.0185173493214958\\
75	33	-0.0156118257312406	-0.0156118257312406\\
75	37	-0.0187571610350563	-0.0187571610350563\\
75	41	0	0\\
85	5	-0.00526167063031413	-0.00526167063031413\\
85	9	-0.022117079627638	-0.022117079627638\\
85	13	-0.0147448976262616	-0.0147448976262616\\
85	17	-0.0236045330455744	-0.0236045330455744\\
85	21	-0.0199566365395513	-0.0199566365395513\\
85	25	-0.00836453583362662	-0.00836453583362662\\
85	29	-0.0141218130230431	-0.0141218130230431\\
85	33	-0.036042686120394	-0.036042686120394\\
85	37	-0.0329795710562457	-0.0329795710562457\\
85	41	-0.00410235306222329	-0.00410235306222329\\
95	5	-0.0117868510931743	-0.0117868510931743\\
95	9	-0.0254638826745379	-0.0254638826745379\\
95	13	-0.0210958038052899	-0.0210958038052899\\
95	17	-0.0348508328200151	-0.0348508328200151\\
95	21	-0.0372479981475362	-0.0372479981475362\\
95	25	-0.0714569133595102	-0.0714569133595102\\
95	29	-1.86813443271427e-05	-1.86813443271427e-05\\
95	33	-0.0277727400269108	-0.0277727400269108\\
95	37	-0.016469395562946	-0.016469395562946\\
95	41	-0.00544326308345422	-0.00544326308345422\\
};
\end{axis}
\end{tikzpicture}%
}
} \\
\subfloat[Difference of the results in Figure \protect\ref{res:graph:scalar:im2}\protect\subref{res:graph:scalar:im2_e} and \protect\ref{res:graph:scalar:im2}\protect\subref{res:graph:scalar:im2_f}.]{\label{res:graph:scalar:im3_c}{
%
%
\begin{tikzpicture}

\begin{axis}[%
width=0.851\figurewidth,
height=\figureheight,
at={(0\figurewidth,0\figureheight)},
scale only axis,
every outer x axis line/.append style={black},
every x tick label/.append style={font=\color{black}},
xmode=log,
xmin=0.01,
xmax=100,
xminorticks=true,
tick align=outside,
xlabel={$\varepsilon$},
xmajorgrids,
xminorgrids,
every outer y axis line/.append style={black},
every y tick label/.append style={font=\color{black}},
ymode=log,
ymin=1,
ymax=100000,
yminorticks=true,
ylabel={$\omega_{0}$},
ymajorgrids,
yminorgrids,
xlabel style={yshift=0.1cm},
ylabel style={yshift=0.1cm},
every outer z axis line/.append style={black},
every z tick label/.append style={font=\color{black}},
zmin=-0.06,
zmax=0.01,
zmajorgrids,
view={146.30993247402}{15.501359566937},
axis background/.style={fill=white},
axis x line*=bottom,
axis y line*=left,
axis z line*=left
]

\addplot3[%
surf,
shader=flat corner,draw=black,z buffer=sort,colormap/jet,mesh/rows=5]
table[row sep=crcr, point meta=\thisrow{c}] {%
x	y	z	c\\
0.01	1	-0.0489162567068631	-0.0489162567068631\\
0.01	10	0.0020367267053476	0.0020367267053476\\
0.01	100	-0.0301265148560285	-0.0301265148560285\\
0.01	1000	-0.0119374748153565	-0.0119374748153565\\
0.01	10000	-0.0371392564464144	-0.0371392564464144\\
0.01	100000	-0.0500649503470069	-0.0500649503470069\\
0.1	1	-0.0302710527438215	-0.0302710527438215\\
0.1	10	-0.0117786259027061	-0.0117786259027061\\
0.1	100	0.00272653770218934	0.00272653770218934\\
0.1	1000	-0.0174751309257827	-0.0174751309257827\\
0.1	10000	0.00511165220794663	0.00511165220794663\\
0.1	100000	5.06432418526659e-05	5.06432418526659e-05\\
1	1	0.00650105384305033	0.00650105384305033\\
1	10	-0.0128079149873486	-0.0128079149873486\\
1	100	0.00257923297418172	0.00257923297418172\\
1	1000	0.00424730080051994	0.00424730080051994\\
1	10000	-0.0171864472826179	-0.0171864472826179\\
1	100000	0.00855166084210635	0.00855166084210635\\
10	1	0	0\\
10	10	0.000478916972131871	0.000478916972131871\\
10	100	-0.0107453674777639	-0.0107453674777639\\
10	1000	0	0\\
10	10000	9.4299630617023e-05	9.4299630617023e-05\\
10	100000	0	0\\
100	1	-0.0119860265217803	-0.0119860265217803\\
100	10	0	0\\
100	100	0	0\\
100	1000	1.93437483068104e-05	1.93437483068104e-05\\
100	10000	-0.0151383649320183	-0.0151383649320183\\
100	100000	0	0\\
};
\end{axis}
\end{tikzpicture}%
}
}
\caption{The differences between the mean for the smooth and non-smooth model with respect to the results in Figure \protect\ref{res:graph:scalar:im2}. Negative values indicate that the non-smooth potential performed better.}
\label{res:graph:scalar:im3}
\end{figure}
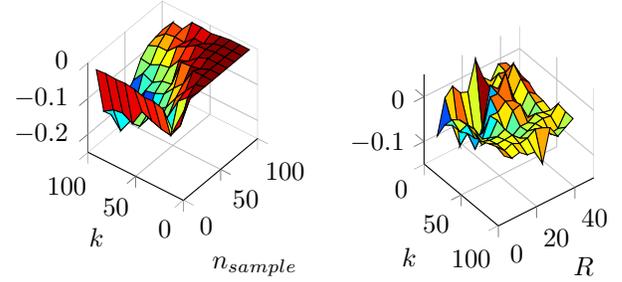
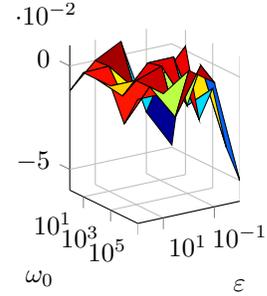

%
Next, we show the effect of varying different parameters. Each plot in Figure \ref{res:graph:scalar:im2} shows the mean of the FOC values which were calculated for $10$ runs with randomly chosen samples. 
In Figure \ref{res:graph:scalar:im2}\subref{res:graph:scalar:im2_a} and \ref{res:graph:scalar:im2}\subref{res:graph:scalar:im2_b}, we vary the number of given sample points $n_{sample}$ and the number of eigenvalues $k$ for the smooth and non-smooth model. For both models, the segmentation performance increases as $n_{sample}$ increases. We observe in the smooth case that if we reduce $n_{sample}$, then $k$ should be reduced as well. 
This effect occurs in the non-smooth case only for the lower range of $n_{sample}$. 
The difference of both results is illustrated in Figure \ref{res:graph:scalar:im3}\subref{res:graph:scalar:im3_a}. Negative values indicate that the non-smooth potential performed better. Except for the case of small values of $k$ and large values of $n_{sample}$, the non-smooth model outperforms the smooth one. 
In Figure \ref{res:graph:scalar:im2}\subref{res:graph:scalar:im2_c} and \ref{res:graph:scalar:im2}\subref{res:graph:scalar:im2_d}, we vary the number of eigenvalues $k$ and the distance $R$ which was a local scale for the graph Laplacian computation, for the smooth and non-smooth model. For both models, the segmentation performance increases as $R$ increases. Small values of $k$ give better results with the non-smooth model. The difference of both results is illustrated in Figure \ref{res:graph:scalar:im3}\subref{res:graph:scalar:im3_b}. In almost every case, the non-smooth model outperforms the smooth one. 
In Figure \ref{res:graph:scalar:im2}\subref{res:graph:scalar:im2_e} and \ref{res:graph:scalar:im2}\subref{res:graph:scalar:im2_f}, we vary the interface parameter $\varepsilon$ and the fidelity parameter $\omega_{0}$ for the smooth and non-smooth model. For both models, the segmentation performance increases as $\varepsilon$ decreases. The difference of both results is illustrated in Figure \ref{res:graph:scalar:im3}\subref{res:graph:scalar:im3_c}. In most cases, the non-smooth model outperforms the smooth one. 

\begin{figure}
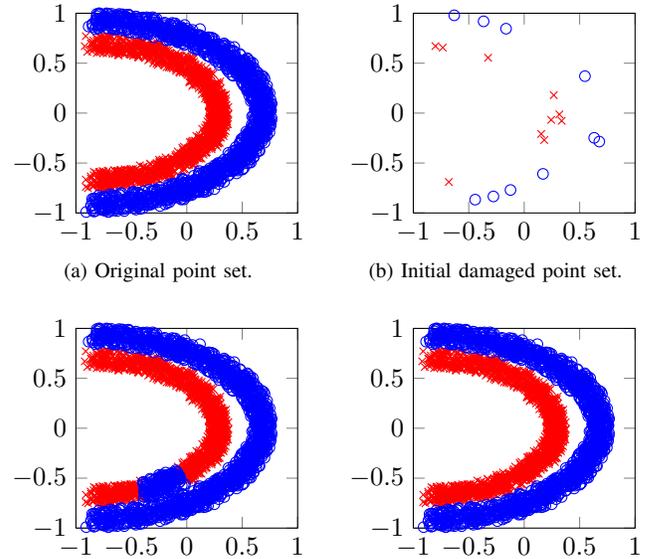

\centering
	\setlength\figureheight{0.3\linewidth} 
	\setlength\figurewidth{0.35\linewidth}
 \subfloat[Original point set.]{\label{res:graph:scalar:im4_a}{
\input{Plots/orig_im_moon.tikz}}
} \ 
\subfloat[Initial damaged point set.]{\label{res:graph:scalar:im4_b}{
%
%
\begin{tikzpicture}

\begin{axis}[%
width=0.951\figurewidth,
height=\figureheight,
at={(0\figurewidth,0\figureheight)},
scale only axis,
xmin=-1,
xmax=1,
ymin=-1,
ymax=1,
axis background/.style={fill=white}
]
\addplot [color=red,only marks,mark=x,mark options={solid},forget plot]
  table[row sep=crcr]{%
-0.734532329264213	0.65614367077679\\
0.243745853418533	-0.0674807170583441\\
0.266569153969524	0.179802523694273\\
0.181064537694652	-0.269781720160618\\
-0.800747770849283	0.671270075983678\\
-0.680121809150042	-0.689874964209221\\
0.153402670315568	-0.20852061305652\\
-0.325742022209366	0.554713259223186\\
0.33782562969333	-0.0777261629990996\\
0.317344379767952	-0.0126467768632227\\
};
\addplot [color=blue,only marks,mark=o,mark options={solid},forget plot]
  table[row sep=crcr]{%
-0.442323990971963	-0.86652815530324\\
-0.366220347463068	0.918865313336691\\
0.169577706569851	-0.608807526893396\\
0.547901209006677	0.370033729788633\\
0.677183743614722	-0.285097215387562\\
-0.631136261377256	0.979348995659024\\
-0.163754624303234	0.845339001191878\\
0.631184506098274	-0.247552937165176\\
-0.12359816871122	-0.770134149490099\\
-0.277730560957741	-0.834556738743438\\
};
\end{axis}
\end{tikzpicture}%
}
}\\
\subfloat[Final segmentation using the smooth model.]{\label{res:graph:scalar:im4_c}{
\input{Plots/smooth_segm_moon.tikz}}
} \ 
\subfloat[Final segmentation using the non-smooth model.]{\label{res:graph:scalar:im4_d}{
\input{Plots/nonsmooth_segm_moon.tikz}}
} 
\caption{Segmentation of a point set into two classes.}
\label{res:graph:scalar:im4}
\end{figure}

Next, we consider a problem with a point set. If not mentioned otherwise, we use the same parameters and stopping criterion as in the previous example. The test is based on the point set given in Figure \ref{res:graph:scalar:im4}\subref{res:graph:scalar:im4_a}, which consists of $3000$ data points in total. 
We have two kinds of points, the red ones and the blue ones, whereby each class contains $1500$ points. 
The damaged data set used as initial state for the smooth and non-smooth model is shown in Figure \ref{res:graph:scalar:im4}\subref{res:graph:scalar:im4_b}. 
The known information is given by $10$ data points for each class. 
Hence, the known data information constitutes only of $0.6667\,\%$ of the whole data set. 
The final segmentation using the smooth and non-smooth model are presented in Figure \ref{res:graph:scalar:im4}\subref{res:graph:scalar:im4_c} and \ref{res:graph:scalar:im4}\subref{res:graph:scalar:im4_d}, 
respectively. 
The chosen parameters are given as $\omega_{0} = 1$, $\eps = 0.5$, $\tau = 0.01$, $\nu=10^{-7}$, $c=3\epsilon^{-1}+\omega_{0}$, 
$R=9$, $k=15$, and $t_{\max}=400$.
The smooth model was not able to correctly classify the area around 
$(-0.75,-0.6)$. This is exactly the area of a large gap in the initial data, 
as seen in Figure \ref{res:graph:scalar:im4}\subref{res:graph:scalar:im4_b}.\\

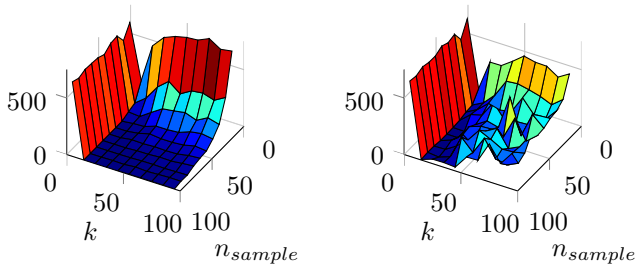
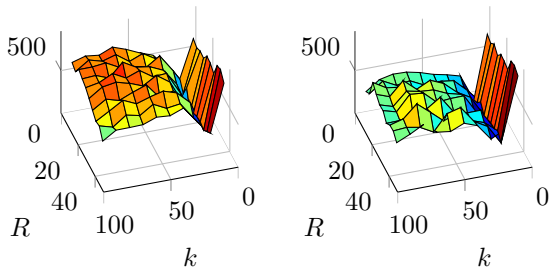
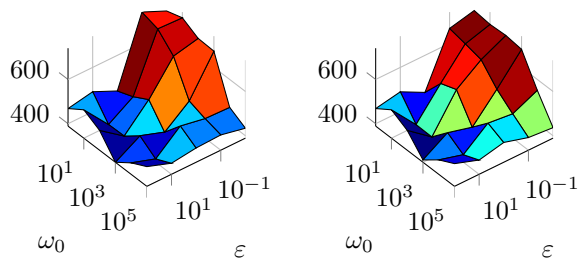
\begin{figure}
\centering
	\setlength\figureheight{0.28\linewidth} 
	\setlength\figurewidth{0.28\linewidth}
 \subfloat[Mean of the misclassification using the smooth model.]{\label{res:graph:scalar:im5_a}{
%
%
\begin{tikzpicture}

\begin{axis}[%
width=0.951\figurewidth,
height=\figureheight,
at={(0\figurewidth,0\figureheight)},
scale only axis,
point meta min=0,
point meta max=751.4,
every outer x axis line/.append style={black},
every x tick label/.append style={font=\color{black}},
xmin=0,
xmax=100,
tick align=outside,
xlabel={$n_{sample}$},
xmajorgrids,
every outer y axis line/.append style={black},
every y tick label/.append style={font=\color{black}},
ymin=0,
ymax=100,
ylabel={$k$},
ymajorgrids,
xlabel style={xshift=-0.4cm,yshift=0.1cm},
ylabel style={yshift=0.2cm,yshift=0.1cm},
every outer z axis line/.append style={black},
every z tick label/.append style={font=\color{black}},
zmin=0,
zmax=751.4,
zmajorgrids,
view={119.357753542791}{41.0746101538247},
axis background/.style={fill=white},
axis x line*=bottom,
axis y line*=left,
axis z line*=left
]

\addplot3[%
surf,
shader=flat corner,draw=black,z buffer=sort,colormap/jet,mesh/rows=10]
table[row sep=crcr, point meta=\thisrow{c}] {%
x	y	z	c\\
10	5	669	669\\
10	15	187.8	187.8\\
10	25	526.9	526.9\\
10	35	632.1	632.1\\
10	45	673.8	673.8\\
10	55	697.5	697.5\\
10	65	706.4	706.4\\
10	75	751.4	751.4\\
10	85	689	689\\
10	95	705.5	705.5\\
20	5	566	566\\
20	15	46.6	46.6\\
20	25	123.2	123.2\\
20	35	253.1	253.1\\
20	45	334.8	334.8\\
20	55	256.5	256.5\\
20	65	314.7	314.7\\
20	75	327.3	327.3\\
20	85	296.5	296.5\\
20	95	353.2	353.2\\
30	5	662.4	662.4\\
30	15	25.9	25.9\\
30	25	17	17\\
30	35	84.7	84.7\\
30	45	149.4	149.4\\
30	55	162.9	162.9\\
30	65	180	180\\
30	75	144.4	144.4\\
30	85	198.8	198.8\\
30	95	172	172\\
40	5	656	656\\
40	15	0.5	0.5\\
40	25	12.3	12.3\\
40	35	17.4	17.4\\
40	45	53.4	53.4\\
40	55	54	54\\
40	65	62	62\\
40	75	100	100\\
40	85	116.1	116.1\\
40	95	100.5	100.5\\
50	5	614.4	614.4\\
50	15	0	0\\
50	25	0	0\\
50	35	14.4	14.4\\
50	45	42.3	42.3\\
50	55	31.3	31.3\\
50	65	31.2	31.2\\
50	75	39.6	39.6\\
50	85	63.4	63.4\\
50	95	51.7	51.7\\
60	5	642.2	642.2\\
60	15	0	0\\
60	25	0	0\\
60	35	15	15\\
60	45	12.2	12.2\\
60	55	17.9	17.9\\
60	65	14.8	14.8\\
60	75	18	18\\
60	85	36.9	36.9\\
60	95	44.9	44.9\\
70	5	649.3	649.3\\
70	15	0	0\\
70	25	0	0\\
70	35	0	0\\
70	45	6.5	6.5\\
70	55	7.4	7.4\\
70	65	14.5	14.5\\
70	75	12.9	12.9\\
70	85	24.5	24.5\\
70	95	33.1	33.1\\
80	5	654.7	654.7\\
80	15	0	0\\
80	25	0	0\\
80	35	0	0\\
80	45	1.8	1.8\\
80	55	5.6	5.6\\
80	65	6.7	6.7\\
80	75	3.9	3.9\\
80	85	7.1	7.1\\
80	95	9.5	9.5\\
90	5	634.1	634.1\\
90	15	0	0\\
90	25	0	0\\
90	35	2.5	2.5\\
90	45	0	0\\
90	55	9.5	9.5\\
90	65	3.1	3.1\\
90	75	8.2	8.2\\
90	85	12.9	12.9\\
90	95	11.6	11.6\\
100	5	658.8	658.8\\
100	15	0.2	0.2\\
100	25	0	0\\
100	35	0	0\\
100	45	0.8	0.8\\
100	55	3.2	3.2\\
100	65	4	4\\
100	75	3.2	3.2\\
100	85	2.7	2.7\\
100	95	1.4	1.4\\
};
\end{axis}
\end{tikzpicture}%
}
} \ 
\subfloat[Mean of the misclassification using the non-smooth model.]{\label{res:graph:scalar:im5_b}{
%
%
\begin{tikzpicture}

\begin{axis}[%
width=0.951\figurewidth,
height=\figureheight,
at={(0\figurewidth,0\figureheight)},
scale only axis,
point meta min=0,
point meta max=751.4,
every outer x axis line/.append style={black},
every x tick label/.append style={font=\color{black}},
xmin=0,
xmax=100,
tick align=outside,
xlabel={$n_{sample}$},
xmajorgrids,
every outer y axis line/.append style={black},
every y tick label/.append style={font=\color{black}},
ymin=0,
ymax=100,
ylabel={$k$},
ymajorgrids,
xlabel style={xshift=-0.4cm,yshift=0.1cm},
ylabel style={yshift=0.2cm,yshift=0.1cm},
every outer z axis line/.append style={black},
every z tick label/.append style={font=\color{black}},
zmin=0,
zmax=751.4,
zmajorgrids,
view={119.357753542791}{41.0746101538247},
axis background/.style={fill=white},
axis x line*=bottom,
axis y line*=left,
axis z line*=left
]

\addplot3[%
surf,
shader=flat corner,draw=black,z buffer=sort,colormap/jet,mesh/rows=10]
table[row sep=crcr, point meta=\thisrow{c}] {%
x	y	z	c\\
10	5	701.8	701.8\\
10	15	117.6	117.6\\
10	25	382.9	382.9\\
10	35	384.5	384.5\\
10	45	455	455\\
10	55	532.8	532.8\\
10	65	510.3	510.3\\
10	75	507.5	507.5\\
10	85	443.2	443.2\\
10	95	503.8	503.8\\
20	5	575.7	575.7\\
20	15	29.1	29.1\\
20	25	81.5	81.5\\
20	35	181	181\\
20	45	335.8	335.8\\
20	55	346.1	346.1\\
20	65	325	325\\
20	75	298.2	298.2\\
20	85	256.7	256.7\\
20	95	335.4	335.4\\
30	5	678.2	678.2\\
30	15	22.2	22.2\\
30	25	77.5	77.5\\
30	35	164.7	164.7\\
30	45	138.2	138.2\\
30	55	156.2	156.2\\
30	65	311.3	311.3\\
30	75	193.8	193.8\\
30	85	182.2	182.2\\
30	95	292.2	292.2\\
40	5	661.4	661.4\\
40	15	7.5	7.5\\
40	25	91	91\\
40	35	119.4	119.4\\
40	45	262.4	262.4\\
40	55	133	133\\
40	65	158	158\\
40	75	163.7	163.7\\
40	85	274.7	274.7\\
40	95	281.7	281.7\\
50	5	688.6	688.6\\
50	15	1.7	1.7\\
50	25	0	0\\
50	35	92.5	92.5\\
50	45	156.7	156.7\\
50	55	131.3	131.3\\
50	65	432.9	432.9\\
50	75	121.8	121.8\\
50	85	341.9	341.9\\
50	95	252.5	252.5\\
60	5	658	658\\
60	15	0	0\\
60	25	0	0\\
60	35	130.9	130.9\\
60	45	166	166\\
60	55	171.7	171.7\\
60	65	136.9	136.9\\
60	75	333.8	333.8\\
60	85	163	163\\
60	95	245.5	245.5\\
70	5	660.3	660.3\\
70	15	0	0\\
70	25	36.6	36.6\\
70	35	118.3	118.3\\
70	45	43.1	43.1\\
70	55	212.3	212.3\\
70	65	269.7	269.7\\
70	75	123.5	123.5\\
70	85	171.1	171.1\\
70	95	311.4	311.4\\
80	5	676.6	676.6\\
80	15	8.9	8.9\\
80	25	12	12\\
80	35	74.7	74.7\\
80	45	158.9	158.9\\
80	55	277.4	277.4\\
80	65	229.3	229.3\\
80	75	175.1	175.1\\
80	85	181	181\\
80	95	204.7	204.7\\
90	5	653.7	653.7\\
90	15	11.1	11.1\\
90	25	0	0\\
90	35	36.5	36.5\\
90	45	240.7	240.7\\
90	55	138.8	138.8\\
90	65	296.7	296.7\\
90	75	145	145\\
90	85	125.6	125.6\\
90	95	171.3	171.3\\
100	5	666	666\\
100	15	5	5\\
100	25	39.6	39.6\\
100	35	41.4	41.4\\
100	45	34	34\\
100	55	154.9	154.9\\
100	65	158.4	158.4\\
100	75	335.2	335.2\\
100	85	194.1	194.1\\
100	95	315.2	315.2\\
};
\end{axis}
\end{tikzpicture}%
}
}\\
\subfloat[Mean of the misclassification using the smooth model.]{\label{res:graph:scalar:im5_c}{
%
%
\begin{tikzpicture}

\begin{axis}[%
width=0.951\figurewidth,
height=\figureheight,
at={(0\figurewidth,0\figureheight)},
scale only axis,
point meta min=65.6,
point meta max=901.4,
every outer x axis line/.append style={black},
every x tick label/.append style={font=\color{black}},
xmin=0,
xmax=100,
tick align=outside,
xlabel={$k$},
xmajorgrids,
every outer y axis line/.append style={black},
every y tick label/.append style={font=\color{black}},
ymin=0,
ymax=50,
ylabel={$R$},
ymajorgrids,
every outer z axis line/.append style={black},
every z tick label/.append style={font=\color{black}},
zmin=65.6,
zmax=901.4,
zmajorgrids,
view={162.474431626277}{45.1078342185956},
axis background/.style={fill=white},
axis x line*=bottom,
axis y line*=left,
axis z line*=left
]

\addplot3[%
surf,
shader=flat corner,draw=black,z buffer=sort,colormap/jet,mesh/rows=10]
table[row sep=crcr, point meta=\thisrow{c}] {%
x	y	z	c\\
5	5	664.7	664.7\\
5	9	634.9	634.9\\
5	13	666.4	666.4\\
5	17	658.5	658.5\\
5	21	741.8	741.8\\
5	25	734.4	734.4\\
5	29	693.9	693.9\\
5	33	840.1	840.1\\
5	37	861.2	861.2\\
5	41	789.6	789.6\\
15	5	281.3	281.3\\
15	9	194.8	194.8\\
15	13	316.8	316.8\\
15	17	131.6	131.6\\
15	21	187.7	187.7\\
15	25	218.5	218.5\\
15	29	228.5	228.5\\
15	33	207.1	207.1\\
15	37	257.9	257.9\\
15	41	305.8	305.8\\
25	5	436.9	436.9\\
25	9	459.8	459.8\\
25	13	515.3	515.3\\
25	17	446	446\\
25	21	423.1	423.1\\
25	25	375	375\\
25	29	402.5	402.5\\
25	33	513.4	513.4\\
25	37	420.5	420.5\\
25	41	451.6	451.6\\
35	5	571.9	571.9\\
35	9	596.5	596.5\\
35	13	693.1	693.1\\
35	17	611.7	611.7\\
35	21	643.5	643.5\\
35	25	618.8	618.8\\
35	29	593.1	593.1\\
35	33	637.8	637.8\\
35	37	589.9	589.9\\
35	41	611.6	611.6\\
45	5	621.6	621.6\\
45	9	678.2	678.2\\
45	13	601.9	601.9\\
45	17	672.7	672.7\\
45	21	633.9	633.9\\
45	25	712.1	712.1\\
45	29	549.3	549.3\\
45	33	680.5	680.5\\
45	37	570.7	570.7\\
45	41	559.1	559.1\\
55	5	682.4	682.4\\
55	9	622.5	622.5\\
55	13	637.6	637.6\\
55	17	693.2	693.2\\
55	21	723.2	723.2\\
55	25	630.1	630.1\\
55	29	705.3	705.3\\
55	33	646.7	646.7\\
55	37	637.9	637.9\\
55	41	569.8	569.8\\
65	5	730.6	730.6\\
65	9	717.1	717.1\\
65	13	652.7	652.7\\
65	17	769.1	769.1\\
65	21	672.3	672.3\\
65	25	683.4	683.4\\
65	29	656.1	656.1\\
65	33	603.3	603.3\\
65	37	547.5	547.5\\
65	41	519.4	519.4\\
75	5	673.1	673.1\\
75	9	675.9	675.9\\
75	13	723.3	723.3\\
75	17	632.7	632.7\\
75	21	742.9	742.9\\
75	25	720.1	720.1\\
75	29	598.4	598.4\\
75	33	664.3	664.3\\
75	37	579.4	579.4\\
75	41	515.2	515.2\\
85	5	690.1	690.1\\
85	9	713.5	713.5\\
85	13	707.3	707.3\\
85	17	684.9	684.9\\
85	21	676.3	676.3\\
85	25	697.3	697.3\\
85	29	637.3	637.3\\
85	33	580.8	580.8\\
85	37	480.3	480.3\\
85	41	513.8	513.8\\
95	5	650.2	650.2\\
95	9	620.1	620.1\\
95	13	706	706\\
95	17	675.9	675.9\\
95	21	650.2	650.2\\
95	25	604.5	604.5\\
95	29	548.9	548.9\\
95	33	501.4	501.4\\
95	37	518	518\\
95	41	427.2	427.2\\
};
\end{axis}
\end{tikzpicture}%
}
} \ 
\subfloat[Mean of the misclassification using the non-smooth model.]{\label{res:graph:scalar:im5_d}{
%
%
\begin{tikzpicture}

\begin{axis}[%
width=0.951\figurewidth,
height=\figureheight,
at={(0\figurewidth,0\figureheight)},
scale only axis,
point meta min=65.6,
point meta max=901.4,
every outer x axis line/.append style={black},
every x tick label/.append style={font=\color{black}},
xmin=0,
xmax=100,
tick align=outside,
xlabel={$k$},
xmajorgrids,
every outer y axis line/.append style={black},
every y tick label/.append style={font=\color{black}},
ymin=0,
ymax=50,
ylabel={$R$},
ymajorgrids,
every outer z axis line/.append style={black},
every z tick label/.append style={font=\color{black}},
zmin=65.6,
zmax=901.4,
zmajorgrids,
view={162.474431626277}{45.1078342185956},
axis background/.style={fill=white},
axis x line*=bottom,
axis y line*=left,
axis z line*=left
]

\addplot3[%
surf,
shader=flat corner,draw=black,z buffer=sort,colormap/jet,mesh/rows=10]
table[row sep=crcr, point meta=\thisrow{c}] {%
x	y	z	c\\
5	5	710	710\\
5	9	681.5	681.5\\
5	13	737.4	737.4\\
5	17	685.1	685.1\\
5	21	759.7	759.7\\
5	25	753.3	753.3\\
5	29	734.2	734.2\\
5	33	885.8	885.8\\
5	37	901.4	901.4\\
5	41	863.3	863.3\\
15	5	65.6	65.6\\
15	9	101.9	101.9\\
15	13	174.5	174.5\\
15	17	99.6	99.6\\
15	21	126	126\\
15	25	151.4	151.4\\
15	29	159	159\\
15	33	203.4	203.4\\
15	37	142.9	142.9\\
15	41	340.1	340.1\\
25	5	334.4	334.4\\
25	9	283.6	283.6\\
25	13	317.5	317.5\\
25	17	309.2	309.2\\
25	21	241	241\\
25	25	274.6	274.6\\
25	29	371.7	371.7\\
25	33	310.1	310.1\\
25	37	334.4	334.4\\
25	41	270.4	270.4\\
35	5	372.6	372.6\\
35	9	342.2	342.2\\
35	13	440.7	440.7\\
35	17	386.6	386.6\\
35	21	429.9	429.9\\
35	25	479.3	479.3\\
35	29	369.8	369.8\\
35	33	356.6	356.6\\
35	37	410.8	410.8\\
35	41	432.9	432.9\\
45	5	401.8	401.8\\
45	9	384.2	384.2\\
45	13	458.9	458.9\\
45	17	501.9	501.9\\
45	21	465.4	465.4\\
45	25	476.9	476.9\\
45	29	401.9	401.9\\
45	33	463.4	463.4\\
45	37	577.8	577.8\\
45	41	482.3	482.3\\
55	5	446	446\\
55	9	385	385\\
55	13	403.5	403.5\\
55	17	559.9	559.9\\
55	21	525.8	525.8\\
55	25	463.4	463.4\\
55	29	579.8	579.8\\
55	33	402.6	402.6\\
55	37	541.6	541.6\\
55	41	424.1	424.1\\
65	5	475.8	475.8\\
65	9	451.2	451.2\\
65	13	423.3	423.3\\
65	17	493.3	493.3\\
65	21	536	536\\
65	25	583.5	583.5\\
65	29	487.7	487.7\\
65	33	517.4	517.4\\
65	37	379.7	379.7\\
65	41	442.7	442.7\\
75	5	480.4	480.4\\
75	9	430.1	430.1\\
75	13	475.3	475.3\\
75	17	607.4	607.4\\
75	21	564.2	564.2\\
75	25	598.1	598.1\\
75	29	443.6	443.6\\
75	33	626.8	626.8\\
75	37	452.5	452.5\\
75	41	546.6	546.6\\
85	5	450.5	450.5\\
85	9	499.3	499.3\\
85	13	495.1	495.1\\
85	17	469.7	469.7\\
85	21	432.9	432.9\\
85	25	442.7	442.7\\
85	29	419.3	419.3\\
85	33	499.5	499.5\\
85	37	477.6	477.6\\
85	41	473	473\\
95	5	353.5	353.5\\
95	9	396.2	396.2\\
95	13	574	574\\
95	17	518.9	518.9\\
95	21	474.8	474.8\\
95	25	480.6	480.6\\
95	29	415.3	415.3\\
95	33	461.4	461.4\\
95	37	467	467\\
95	41	411.3	411.3\\
};
\end{axis}
\end{tikzpicture}%
}
}\\
\subfloat[Mean of the misclassification using the smooth model.]{\label{res:graph:scalar:im5_e}{
%
%
\begin{tikzpicture}

\begin{axis}[%
width=0.951\figurewidth,
height=\figureheight,
at={(0\figurewidth,0\figureheight)},
scale only axis,
every outer x axis line/.append style={black},
every x tick label/.append style={font=\color{black}},
xmode=log,
xmin=0.01,
xmax=100,
xminorticks=true,
tick align=outside,
xlabel={$\varepsilon$},
xmajorgrids,
xminorgrids,
every outer y axis line/.append style={black},
every y tick label/.append style={font=\color{black}},
ymode=log,
ymin=1,
ymax=100000,
yminorticks=true,
ylabel={$\omega_{0}$},
ymajorgrids,
yminorgrids,
every outer z axis line/.append style={black},
every z tick label/.append style={font=\color{black}},
zmin=389.1,
zmax=741.2,
zmajorgrids,
view={141.34019174591}{44.686044238781},
axis background/.style={fill=white},
axis x line*=bottom,
axis y line*=left,
axis z line*=left
]

\addplot3[%
surf,
shader=flat corner,draw=black,z buffer=sort,colormap/jet,mesh/rows=5]
table[row sep=crcr, point meta=\thisrow{c}] {%
x	y	z	c\\
0.01	1	691.7	691.7\\
0.01	10	715.3	715.3\\
0.01	100	676.7	676.7\\
0.01	1000	673.4	673.4\\
0.01	10000	473.8	473.8\\
0.01	100000	434	434\\
0.1	1	741.2	741.2\\
0.1	10	720.1	720.1\\
0.1	100	651.4	651.4\\
0.1	1000	503.2	503.2\\
0.1	10000	477.7	477.7\\
0.1	100000	458.1	458.1\\
1	1	439.1	439.1\\
1	10	467.8	467.8\\
1	100	509	509\\
1	1000	416.4	416.4\\
1	10000	498.8	498.8\\
1	100000	453.7	453.7\\
10	1	490	490\\
10	10	396.6	396.6\\
10	100	398.6	398.6\\
10	1000	454.5	454.5\\
10	10000	404.2	404.2\\
10	100000	439.1	439.1\\
100	1	468.4	468.4\\
100	10	517.3	517.3\\
100	100	478.6	478.6\\
100	1000	389.1	389.1\\
100	10000	440.8	440.8\\
100	100000	478.2	478.2\\
};
\end{axis}
\end{tikzpicture}%
}
} \ 
\subfloat[Mean of the misclassification using the non-smooth model.]{\label{res:graph:scalar:im5_f}{
%
%
\begin{tikzpicture}

\begin{axis}[%
width=0.951\figurewidth,
height=\figureheight,
at={(0\figurewidth,0\figureheight)},
scale only axis,
every outer x axis line/.append style={black},
every x tick label/.append style={font=\color{black}},
xmode=log,
xmin=0.01,
xmax=100,
xminorticks=true,
tick align=outside,
xlabel={$\varepsilon$},
xmajorgrids,
xminorgrids,
every outer y axis line/.append style={black},
every y tick label/.append style={font=\color{black}},
ymode=log,
ymin=1,
ymax=100000,
yminorticks=true,
ylabel={$\omega_{0}$},
ymajorgrids,
yminorgrids,
every outer z axis line/.append style={black},
every z tick label/.append style={font=\color{black}},
zmin=389.1,
zmax=741.2,
zmajorgrids,
view={141.34019174591}{44.686044238781},
axis background/.style={fill=white},
axis x line*=bottom,
axis y line*=left,
axis z line*=left
]

\addplot3[%
surf,
shader=flat corner,draw=black,z buffer=sort,colormap/jet,mesh/rows=5]
table[row sep=crcr, point meta=\thisrow{c}] {%
x	y	z	c\\
0.01	1	693.7	693.7\\
0.01	10	685.6	685.6\\
0.01	100	690.3	690.3\\
0.01	1000	681.7	681.7\\
0.01	10000	551.1	551.1\\
0.01	100000	436.6	436.6\\
0.1	1	667.3	667.3\\
0.1	10	652.3	652.3\\
0.1	100	635.4	635.4\\
0.1	1000	556.4	556.4\\
0.1	10000	499.3	499.3\\
0.1	100000	446.3	446.3\\
1	1	432.6	432.6\\
1	10	527.5	527.5\\
1	100	554.6	554.6\\
1	1000	438.9	438.9\\
1	10000	511.8	511.8\\
1	100000	478.6	478.6\\
10	1	489.7	489.7\\
10	10	396.1	396.1\\
10	100	394.9	394.9\\
10	1000	468.5	468.5\\
10	10000	423.2	423.2\\
10	100000	453.5	453.5\\
100	1	468.4	468.4\\
100	10	517.3	517.3\\
100	100	478.9	478.9\\
100	1000	399.6	399.6\\
100	10000	454.2	454.2\\
100	100000	491.7	491.7\\
};
\end{axis}
\end{tikzpicture}%
}
}
\caption{Comparison of the smooth and non-smooth model: The mean of the misclassification for the smooth (left column) and non-smooth (right column) model for varying parameters. For each $(x,y)$ pair, we have taken $10$ runs with randomly chosen samples.}
\label{res:graph:scalar:im5}
\end{figure}

\begin{figure}
\centering
	\setlength\figureheight{0.3\linewidth} 
	\setlength\figurewidth{0.3\linewidth}
\subfloat[Difference of the results in Figure \protect\ref{res:graph:scalar:im5}\protect\subref{res:graph:scalar:im5_a} and \protect\ref{res:graph:scalar:im5}\protect\subref{res:graph:scalar:im5_b}.]{\label{res:graph:scalar:im6_a}{
%
%
\begin{tikzpicture}

\begin{axis}[%
width=0.851\figurewidth,
height=\figureheight,
at={(0\figurewidth,0\figureheight)},
scale only axis,
every outer x axis line/.append style={black},
every x tick label/.append style={font=\color{black}},
xmin=0,
xmax=100,
tick align=outside,
xlabel={$n_{sample}$},
xmajorgrids,
every outer y axis line/.append style={black},
every y tick label/.append style={font=\color{black}},
ymin=0,
ymax=100,
ylabel={$k$},
ymajorgrids,
xlabel style={xshift=-0.4cm,yshift=0.1cm},
ylabel style={yshift=0.2cm,yshift=0.1cm},
every outer z axis line/.append style={black},
every z tick label/.append style={font=\color{black}},
zmin=-400,
zmax=600,
zmajorgrids,
view={119.357753542791}{41.0746101538247},
axis background/.style={fill=white},
axis x line*=bottom,
axis y line*=left,
axis z line*=left
]

\addplot3[%
surf,
shader=flat corner,draw=black,z buffer=sort,colormap/jet,mesh/rows=10]
table[row sep=crcr, point meta=\thisrow{c}] {%
x	y	z	c\\
10	5	32.8	32.8\\
10	15	-70.2	-70.2\\
10	25	-144	-144\\
10	35	-247.6	-247.6\\
10	45	-218.8	-218.8\\
10	55	-164.7	-164.7\\
10	65	-196.1	-196.1\\
10	75	-243.9	-243.9\\
10	85	-245.8	-245.8\\
10	95	-201.7	-201.7\\
20	5	9.70000000000005	9.70000000000005\\
20	15	-17.5	-17.5\\
20	25	-41.7	-41.7\\
20	35	-72.1	-72.1\\
20	45	1	1\\
20	55	89.6	89.6\\
20	65	10.3	10.3\\
20	75	-29.1	-29.1\\
20	85	-39.8	-39.8\\
20	95	-17.8	-17.8\\
30	5	15.8000000000001	15.8000000000001\\
30	15	-3.7	-3.7\\
30	25	60.5	60.5\\
30	35	80	80\\
30	45	-11.2	-11.2\\
30	55	-6.70000000000002	-6.70000000000002\\
30	65	131.3	131.3\\
30	75	49.4	49.4\\
30	85	-16.6	-16.6\\
30	95	120.2	120.2\\
40	5	5.39999999999998	5.39999999999998\\
40	15	7	7\\
40	25	78.7	78.7\\
40	35	102	102\\
40	45	209	209\\
40	55	79	79\\
40	65	96	96\\
40	75	63.7	63.7\\
40	85	158.6	158.6\\
40	95	181.2	181.2\\
50	5	74.2	74.2\\
50	15	1.7	1.7\\
50	25	0	0\\
50	35	78.1	78.1\\
50	45	114.4	114.4\\
50	55	100	100\\
50	65	401.7	401.7\\
50	75	82.2	82.2\\
50	85	278.5	278.5\\
50	95	200.8	200.8\\
60	5	15.8	15.8\\
60	15	0	0\\
60	25	0	0\\
60	35	115.9	115.9\\
60	45	153.8	153.8\\
60	55	153.8	153.8\\
60	65	122.1	122.1\\
60	75	315.8	315.8\\
60	85	126.1	126.1\\
60	95	200.6	200.6\\
70	5	11	11\\
70	15	0	0\\
70	25	36.6	36.6\\
70	35	118.3	118.3\\
70	45	36.6	36.6\\
70	55	204.9	204.9\\
70	65	255.2	255.2\\
70	75	110.6	110.6\\
70	85	146.6	146.6\\
70	95	278.3	278.3\\
80	5	21.9	21.9\\
80	15	8.9	8.9\\
80	25	12	12\\
80	35	74.7	74.7\\
80	45	157.1	157.1\\
80	55	271.8	271.8\\
80	65	222.6	222.6\\
80	75	171.2	171.2\\
80	85	173.9	173.9\\
80	95	195.2	195.2\\
90	5	19.6	19.6\\
90	15	11.1	11.1\\
90	25	0	0\\
90	35	34	34\\
90	45	240.7	240.7\\
90	55	129.3	129.3\\
90	65	293.6	293.6\\
90	75	136.8	136.8\\
90	85	112.7	112.7\\
90	95	159.7	159.7\\
100	5	7.20000000000005	7.20000000000005\\
100	15	4.8	4.8\\
100	25	39.6	39.6\\
100	35	41.4	41.4\\
100	45	33.2	33.2\\
100	55	151.7	151.7\\
100	65	154.4	154.4\\
100	75	332	332\\
100	85	191.4	191.4\\
100	95	313.8	313.8\\
};
\end{axis}
\end{tikzpicture}%
}
} \ 
\subfloat[Difference of the results in Figure \protect\ref{res:graph:scalar:im5}\protect\subref{res:graph:scalar:im5_c} and \protect\ref{res:graph:scalar:im5}\protect\subref{res:graph:scalar:im5_d}.]{\label{res:graph:scalar:im6_b}{
%
%
\begin{tikzpicture}

\begin{axis}[%
width=0.851\figurewidth,
height=\figureheight,
at={(0\figurewidth,0\figureheight)},
scale only axis,
every outer x axis line/.append style={black},
every x tick label/.append style={font=\color{black}},
xmin=0,
xmax=100,
tick align=outside,
xlabel={$k$},
xmajorgrids,
every outer y axis line/.append style={black},
every y tick label/.append style={font=\color{black}},
ymin=0,
ymax=50,
ylabel={$R$},
ymajorgrids,
every outer z axis line/.append style={black},
every z tick label/.append style={font=\color{black}},
zmin=-300,
zmax=100,
zmajorgrids,
view={162.474431626277}{45.1078342185956},
axis background/.style={fill=white},
axis x line*=bottom,
axis y line*=left,
axis z line*=left
]

\addplot3[%
surf,
shader=flat corner,draw=black,z buffer=sort,colormap/jet,mesh/rows=10]
table[row sep=crcr, point meta=\thisrow{c}] {%
x	y	z	c\\
5	5	45.3	45.3\\
5	9	46.6	46.6\\
5	13	71	71\\
5	17	26.6	26.6\\
5	21	17.9000000000001	17.9000000000001\\
5	25	18.9	18.9\\
5	29	40.3000000000001	40.3000000000001\\
5	33	45.6999999999999	45.6999999999999\\
5	37	40.1999999999999	40.1999999999999\\
5	41	73.6999999999999	73.6999999999999\\
15	5	-215.7	-215.7\\
15	9	-92.9	-92.9\\
15	13	-142.3	-142.3\\
15	17	-32	-32\\
15	21	-61.7	-61.7\\
15	25	-67.1	-67.1\\
15	29	-69.5	-69.5\\
15	33	-3.69999999999999	-3.69999999999999\\
15	37	-115	-115\\
15	41	34.3	34.3\\
25	5	-102.5	-102.5\\
25	9	-176.2	-176.2\\
25	13	-197.8	-197.8\\
25	17	-136.8	-136.8\\
25	21	-182.1	-182.1\\
25	25	-100.4	-100.4\\
25	29	-30.8	-30.8\\
25	33	-203.3	-203.3\\
25	37	-86.1	-86.1\\
25	41	-181.2	-181.2\\
35	5	-199.3	-199.3\\
35	9	-254.3	-254.3\\
35	13	-252.4	-252.4\\
35	17	-225.1	-225.1\\
35	21	-213.6	-213.6\\
35	25	-139.5	-139.5\\
35	29	-223.3	-223.3\\
35	33	-281.2	-281.2\\
35	37	-179.1	-179.1\\
35	41	-178.7	-178.7\\
45	5	-219.8	-219.8\\
45	9	-294	-294\\
45	13	-143	-143\\
45	17	-170.8	-170.8\\
45	21	-168.5	-168.5\\
45	25	-235.2	-235.2\\
45	29	-147.4	-147.4\\
45	33	-217.1	-217.1\\
45	37	7.09999999999991	7.09999999999991\\
45	41	-76.8	-76.8\\
55	5	-236.4	-236.4\\
55	9	-237.5	-237.5\\
55	13	-234.1	-234.1\\
55	17	-133.3	-133.3\\
55	21	-197.4	-197.4\\
55	25	-166.7	-166.7\\
55	29	-125.5	-125.5\\
55	33	-244.1	-244.1\\
55	37	-96.3	-96.3\\
55	41	-145.7	-145.7\\
65	5	-254.8	-254.8\\
65	9	-265.9	-265.9\\
65	13	-229.4	-229.4\\
65	17	-275.8	-275.8\\
65	21	-136.3	-136.3\\
65	25	-99.9	-99.9\\
65	29	-168.4	-168.4\\
65	33	-85.9	-85.9\\
65	37	-167.8	-167.8\\
65	41	-76.7	-76.7\\
75	5	-192.7	-192.7\\
75	9	-245.8	-245.8\\
75	13	-248	-248\\
75	17	-25.3000000000001	-25.3000000000001\\
75	21	-178.7	-178.7\\
75	25	-122	-122\\
75	29	-154.8	-154.8\\
75	33	-37.5	-37.5\\
75	37	-126.9	-126.9\\
75	41	31.4	31.4\\
85	5	-239.6	-239.6\\
85	9	-214.2	-214.2\\
85	13	-212.2	-212.2\\
85	17	-215.2	-215.2\\
85	21	-243.4	-243.4\\
85	25	-254.6	-254.6\\
85	29	-218	-218\\
85	33	-81.3	-81.3\\
85	37	-2.69999999999999	-2.69999999999999\\
85	41	-40.8	-40.8\\
95	5	-296.7	-296.7\\
95	9	-223.9	-223.9\\
95	13	-132	-132\\
95	17	-157	-157\\
95	21	-175.4	-175.4\\
95	25	-123.9	-123.9\\
95	29	-133.6	-133.6\\
95	33	-40	-40\\
95	37	-51	-51\\
95	41	-15.9	-15.9\\
};
\end{axis}
\end{tikzpicture}%
}
} \\
\subfloat[Difference of the results in Figure \protect\ref{res:graph:scalar:im5}\protect\subref{res:graph:scalar:im5_e} and \protect\ref{res:graph:scalar:im5}\protect\subref{res:graph:scalar:im5_f}.]{\label{res:graph:scalar:im6_c}{
%
%
\begin{tikzpicture}

\begin{axis}[%
width=0.851\figurewidth,
height=\figureheight,
at={(0\figurewidth,0\figureheight)},
scale only axis,
every outer x axis line/.append style={black},
every x tick label/.append style={font=\color{black}},
xmode=log,
xmin=0.01,
xmax=100,
xminorticks=true,
tick align=outside,
xlabel={$\varepsilon$},
xmajorgrids,
xminorgrids,
every outer y axis line/.append style={black},
every y tick label/.append style={font=\color{black}},
ymode=log,
ymin=1,
ymax=100000,
yminorticks=true,
ylabel={$\omega_{0}$},
ymajorgrids,
yminorgrids,
every outer z axis line/.append style={black},
every z tick label/.append style={font=\color{black}},
zmin=-100,
zmax=100,
zmajorgrids,
view={141.34019174591}{44.686044238781},
axis background/.style={fill=white},
axis x line*=bottom,
axis y line*=left,
axis z line*=left
]

\addplot3[%
surf,
shader=flat corner,draw=black,z buffer=sort,colormap/jet,mesh/rows=5]
table[row sep=crcr, point meta=\thisrow{c}] {%
x	y	z	c\\
0.01	1	2	2\\
0.01	10	-29.6999999999999	-29.6999999999999\\
0.01	100	13.5999999999999	13.5999999999999\\
0.01	1000	8.30000000000007	8.30000000000007\\
0.01	10000	77.3	77.3\\
0.01	100000	2.60000000000002	2.60000000000002\\
0.1	1	-73.9000000000001	-73.9000000000001\\
0.1	10	-67.8000000000001	-67.8000000000001\\
0.1	100	-16	-16\\
0.1	1000	53.2	53.2\\
0.1	10000	21.6	21.6\\
0.1	100000	-11.8	-11.8\\
1	1	-6.5	-6.5\\
1	10	59.7	59.7\\
1	100	45.6	45.6\\
1	1000	22.5	22.5\\
1	10000	13	13\\
1	100000	24.9	24.9\\
10	1	-0.300000000000011	-0.300000000000011\\
10	10	-0.5	-0.5\\
10	100	-3.70000000000005	-3.70000000000005\\
10	1000	14	14\\
10	10000	19	19\\
10	100000	14.4	14.4\\
100	1	0	0\\
100	10	0	0\\
100	100	0.299999999999955	0.299999999999955\\
100	1000	10.5	10.5\\
100	10000	13.4	13.4\\
100	100000	13.5	13.5\\
};
\end{axis}
\end{tikzpicture}%
}
}
\caption{The differences between the mean for the smooth and non-smooth model with respect to the results in Figure \protect\ref{res:graph:scalar:im5}. Negative values indicate that the non-smooth potential performed better.}
\label{res:graph:scalar:im6}
\end{figure}
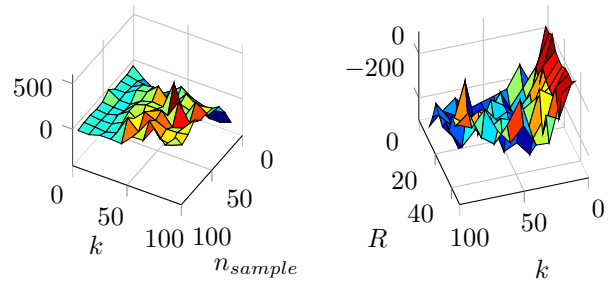
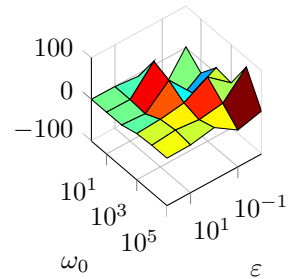

Next, we show the effect of varying different parameters. Each plot in Figure \ref{res:graph:scalar:im5} shows the mean of the number of misclassified points which were calculated for $10$ runs with randomly chosen samples. 
In Figure \ref{res:graph:scalar:im5}\subref{res:graph:scalar:im5_a} and \ref{res:graph:scalar:im5}\subref{res:graph:scalar:im5_b}, we vary the number of given sample points $n_{sample}$ and the number of eigenvalues $k$ for the smooth and non-smooth model. For both models, the segmentation performance increases as $n_{sample}$ increases. 
For small values of $n_{sample}$, the non-smooth model performs better, whereas the smooth model gives better results for larger values of the pair $(n_{sample},k)$. This can be seen in Figure \ref{res:graph:scalar:im6}\subref{res:graph:scalar:im6_a}, which shows the difference of both results. Negative values indicate that the non-smooth potential performed better. 
In Figure \ref{res:graph:scalar:im5}\subref{res:graph:scalar:im5_c} and \ref{res:graph:scalar:im5}\subref{res:graph:scalar:im5_d}, we vary the number of eigenvalues $k$ and the distance $R$ for the smooth and non-smooth model. For both models, the segmentation performance is the best for $k=10$. The difference of both results is illustrated in Figure \ref{res:graph:scalar:im6}\subref{res:graph:scalar:im6_b}. In almost every case, the non-smooth model outperforms the smooth one. 
In Figure \ref{res:graph:scalar:im5}\subref{res:graph:scalar:im5_e} and \ref{res:graph:scalar:im5}\subref{res:graph:scalar:im5_f}, we vary the interface parameter $\varepsilon$ and the fidelity parameter $\omega_{0}$ for the smooth and non-smooth model. For both models, the segmentation performance increases as $\varepsilon$ and $\omega_{0}$ increase. The difference of both results is illustrated in Figure \ref{res:graph:scalar:im6}\subref{res:graph:scalar:im6_c}. Both models behave similar. 

\subsubsection*{Multiclass segmentation}
\begin{figure}
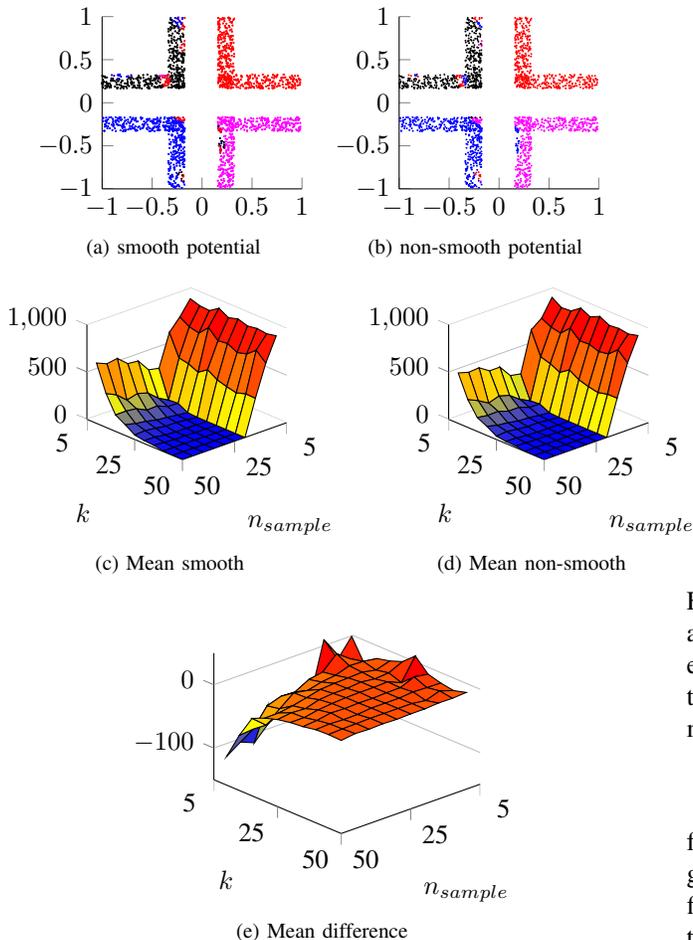
%
\centering
\subfloat[smooth potential]{
\input{Plots/datavv1_s.tikz}
} 
\subfloat[non-smooth potential]{
\input{Plots/datavv1_ns.tikz}
}\\


\subfloat[Mean smooth]{
%
%
\begin{tikzpicture}

\begin{axis}[%
width=.3\linewidth,
at={(0.783in,0.497in)},
scale only axis,
xmin=0,
xmax=10,
xtick={0,5,10},
xticklabels={5,25,50},
ytick={0,5,10},
yticklabels={5,25,50},
tick align=outside,
ymin=0,
ymax=10,
ylabel={$k$},
xlabel={$n_{sample}$},
zmin=0,
zmax=1000,
zmajorgrids,
view={137.5}{30},
axis background/.style={fill=white},
axis x line*=bottom,
axis y line*=left,
axis z line*=left
]

\addplot3[%
surf,
shader=flat corner,draw=black,z buffer=sort,mesh/rows=10]
table[row sep=crcr, point meta=\thisrow{c}] {%
x	y	z	c\\
1	1	963.4	963.4\\
1	2	931.8	931.8\\
1	3	923.8	923.8\\
1	4	990.3	990.3\\
1	5	933.8	933.8\\
1	6	963	963\\
1	7	931.3	931.3\\
1	8	947.3	947.3\\
1	9	934.7	934.7\\
1	10	955.4	955.4\\
2	1	810.6	810.6\\
2	2	749.7	749.7\\
2	3	725.8	725.8\\
2	4	826	826\\
2	5	730.4	730.4\\
2	6	743.3	743.3\\
2	7	691.4	691.4\\
2	8	718	718\\
2	9	800.6	800.6\\
2	10	707.9	707.9\\
3	1	687.5	687.5\\
3	2	460.8	460.8\\
3	3	441	441\\
3	4	393.7	393.7\\
3	5	463.9	463.9\\
3	6	399.2	399.2\\
3	7	369.6	369.6\\
3	8	347.3	347.3\\
3	9	378.4	378.4\\
3	10	374.6	374.6\\
4	1	347.5	347.5\\
4	2	55.4	55.4\\
4	3	62.4	62.4\\
4	4	7.4	7.4\\
4	5	0	0\\
4	6	0	0\\
4	7	0	0\\
4	8	0	0\\
4	9	0	0\\
4	10	0	0\\
5	1	369.5	369.5\\
5	2	114.6	114.6\\
5	3	66.5	66.5\\
5	4	18.8	18.8\\
5	5	10.1	10.1\\
5	6	0	0\\
5	7	1	1\\
5	8	0	0\\
5	9	0	0\\
5	10	0	0\\
6	1	501.5	501.5\\
6	2	208.3	208.3\\
6	3	112.7	112.7\\
6	4	34.5	34.5\\
6	5	23.3	23.3\\
6	6	4.4	4.4\\
6	7	2.7	2.7\\
6	8	0	0\\
6	9	0	0\\
6	10	5.9	5.9\\
7	1	513.3	513.3\\
7	2	239.9	239.9\\
7	3	107.1	107.1\\
7	4	45.2	45.2\\
7	5	22.7	22.7\\
7	6	4.9	4.9\\
7	7	6.6	6.6\\
7	8	4.1	4.1\\
7	9	0	0\\
7	10	0	0\\
8	1	606.9	606.9\\
8	2	211.5	211.5\\
8	3	161.1	161.1\\
8	4	70.7	70.7\\
8	5	41.3	41.3\\
8	6	12.1	12.1\\
8	7	5.9	5.9\\
8	8	4.6	4.6\\
8	9	2.1	2.1\\
8	10	1	1\\
9	1	589.5	589.5\\
9	2	304.5	304.5\\
9	3	175.3	175.3\\
9	4	86.5	86.5\\
9	5	54.5	54.5\\
9	6	28.6	28.6\\
9	7	15.2	15.2\\
9	8	5.5	5.5\\
9	9	1.9	1.9\\
9	10	0	0\\
10	1	636.5	636.5\\
10	2	357.3	357.3\\
10	3	173.6	173.6\\
10	4	114.4	114.4\\
10	5	47.3	47.3\\
10	6	30.4	30.4\\
10	7	23	23\\
10	8	6.2	6.2\\
10	9	3.6	3.6\\
10	10	3.9	3.9\\
};
\end{axis}
\end{tikzpicture}%
} 
\subfloat[Mean non-smooth]{
%
%
\begin{tikzpicture}

\begin{axis}[%
width=.3\linewidth,
at={(0.783in,0.497in)},
scale only axis,
xmin=0,
xmax=10,
xtick={0,5,10},
xticklabels={5,25,50},
ytick={0,5,10},
yticklabels={5,25,50},
tick align=outside,
ymin=0,
ymax=10,
ylabel={$k$},
xlabel={$n_{sample}$},
zmin=0,
zmax=1000,
zmajorgrids,
view={137.5}{30},
axis background/.style={fill=white},
axis x line*=bottom,
axis y line*=left,
axis z line*=left
]

\addplot3[%
surf,
shader=flat corner,draw=black,z buffer=sort,mesh/rows=10]
table[row sep=crcr, point meta=\thisrow{c}] {%
x	y	z	c\\
1	1	978.4	978.4\\
1	2	911.6	911.6\\
1	3	920.6	920.6\\
1	4	991.8	991.8\\
1	5	942.8	942.8\\
1	6	989.4	989.4\\
1	7	937.4	937.4\\
1	8	947.9	947.9\\
1	9	931.9	931.9\\
1	10	958.2	958.2\\
2	1	797.1	797.1\\
2	2	745.3	745.3\\
2	3	725.9	725.9\\
2	4	820.9	820.9\\
2	5	729.3	729.3\\
2	6	735.2	735.2\\
2	7	700.8	700.8\\
2	8	719.1	719.1\\
2	9	805.1	805.1\\
2	10	709.2	709.2\\
3	1	712.5	712.5\\
3	2	472.3	472.3\\
3	3	441.9	441.9\\
3	4	392.9	392.9\\
3	5	461.1	461.1\\
3	6	400.2	400.2\\
3	7	369.9	369.9\\
3	8	349.4	349.4\\
3	9	375.4	375.4\\
3	10	376.2	376.2\\
4	1	287.7	287.7\\
4	2	45.8	45.8\\
4	3	56	56\\
4	4	5.4	5.4\\
4	5	0	0\\
4	6	0	0\\
4	7	0	0\\
4	8	0	0\\
4	9	0	0\\
4	10	0	0\\
5	1	309.1	309.1\\
5	2	100.9	100.9\\
5	3	57.4	57.4\\
5	4	16	16\\
5	5	7.9	7.9\\
5	6	0	0\\
5	7	0	0\\
5	8	0	0\\
5	9	0	0\\
5	10	0	0\\
6	1	437.1	437.1\\
6	2	182.7	182.7\\
6	3	98	98\\
6	4	29.9	29.9\\
6	5	20.9	20.9\\
6	6	3.8	3.8\\
6	7	2.1	2.1\\
6	8	0	0\\
6	9	0	0\\
6	10	5.1	5.1\\
7	1	436.8	436.8\\
7	2	215.3	215.3\\
7	3	96.5	96.5\\
7	4	35.3	35.3\\
7	5	18.2	18.2\\
7	6	2.8	2.8\\
7	7	5	5\\
7	8	1.7	1.7\\
7	9	0	0\\
7	10	0	0\\
8	1	508.2	508.2\\
8	2	179.7	179.7\\
8	3	133.2	133.2\\
8	4	55.8	55.8\\
8	5	30.4	30.4\\
8	6	7.7	7.7\\
8	7	5.2	5.2\\
8	8	1.8	1.8\\
8	9	1.5	1.5\\
8	10	0.1	0.1\\
9	1	499.7	499.7\\
9	2	243.7	243.7\\
9	3	129.4	129.4\\
9	4	68.1	68.1\\
9	5	44.9	44.9\\
9	6	19.9	19.9\\
9	7	10.5	10.5\\
9	8	3.8	3.8\\
9	9	0.8	0.8\\
9	10	0	0\\
10	1	531.6	531.6\\
10	2	294.4	294.4\\
10	3	146	146\\
10	4	95.7	95.7\\
10	5	34.8	34.8\\
10	6	21.7	21.7\\
10	7	16.3	16.3\\
10	8	3.2	3.2\\
10	9	1.4	1.4\\
10	10	0.9	0.9\\
};
\end{axis}
\end{tikzpicture}%
} \\

\subfloat[Mean difference]{
%
%
\begin{tikzpicture}

\begin{axis}[%
width=.4\linewidth,
at={(0.783in,0.497in)},
scale only axis,
xmin=0,
xmax=10,
xtick={0,5,10},
xticklabels={5,25,50},
ytick={0,5,10},
yticklabels={5,25,50},
tick align=outside,
ylabel={$k$},
xlabel={$n_{sample}$},
ymin=0,
ymax=10,
zmin=-150,
zmax=50,
zmajorgrids,
view={137.5}{30},
axis background/.style={fill=white},
axis x line*=bottom,
axis y line*=left,
axis z line*=left
]

\addplot3[%
surf,
shader=flat corner,draw=black,z buffer=sort,mesh/rows=10]
table[row sep=crcr, point meta=\thisrow{c}] {%
x	y	z	c\\
1	1	15	15\\
1	2	-20.1999999999999	-20.1999999999999\\
1	3	-3.19999999999993	-3.19999999999993\\
1	4	1.5	1.5\\
1	5	9	9\\
1	6	26.4	26.4\\
1	7	6.10000000000002	6.10000000000002\\
1	8	0.600000000000023	0.600000000000023\\
1	9	-2.80000000000007	-2.80000000000007\\
1	10	2.80000000000007	2.80000000000007\\
2	1	-13.5	-13.5\\
2	2	-4.40000000000009	-4.40000000000009\\
2	3	0.100000000000023	0.100000000000023\\
2	4	-5.10000000000002	-5.10000000000002\\
2	5	-1.10000000000002	-1.10000000000002\\
2	6	-8.09999999999991	-8.09999999999991\\
2	7	9.39999999999998	9.39999999999998\\
2	8	1.10000000000002	1.10000000000002\\
2	9	4.5	4.5\\
2	10	1.30000000000007	1.30000000000007\\
3	1	25	25\\
3	2	11.5	11.5\\
3	3	0.899999999999977	0.899999999999977\\
3	4	-0.800000000000011	-0.800000000000011\\
3	5	-2.79999999999995	-2.79999999999995\\
3	6	1	1\\
3	7	0.299999999999955	0.299999999999955\\
3	8	2.09999999999997	2.09999999999997\\
3	9	-3	-3\\
3	10	1.59999999999997	1.59999999999997\\
4	1	-59.8	-59.8\\
4	2	-9.6	-9.6\\
4	3	-6.4	-6.4\\
4	4	-2	-2\\
4	5	0	0\\
4	6	0	0\\
4	7	0	0\\
4	8	0	0\\
4	9	0	0\\
4	10	0	0\\
5	1	-60.4	-60.4\\
5	2	-13.7	-13.7\\
5	3	-9.1	-9.1\\
5	4	-2.8	-2.8\\
5	5	-2.2	-2.2\\
5	6	0	0\\
5	7	-1	-1\\
5	8	0	0\\
5	9	0	0\\
5	10	0	0\\
6	1	-64.4	-64.4\\
6	2	-25.6	-25.6\\
6	3	-14.7	-14.7\\
6	4	-4.6	-4.6\\
6	5	-2.4	-2.4\\
6	6	-0.600000000000001	-0.600000000000001\\
6	7	-0.6	-0.6\\
6	8	0	0\\
6	9	0	0\\
6	10	-0.800000000000001	-0.800000000000001\\
7	1	-76.4999999999999	-76.4999999999999\\
7	2	-24.6	-24.6\\
7	3	-10.6	-10.6\\
7	4	-9.90000000000001	-9.90000000000001\\
7	5	-4.5	-4.5\\
7	6	-2.1	-2.1\\
7	7	-1.6	-1.6\\
7	8	-2.4	-2.4\\
7	9	0	0\\
7	10	0	0\\
8	1	-98.7	-98.7\\
8	2	-31.8	-31.8\\
8	3	-27.9	-27.9\\
8	4	-14.9	-14.9\\
8	5	-10.9	-10.9\\
8	6	-4.4	-4.4\\
8	7	-0.7	-0.7\\
8	8	-2.8	-2.8\\
8	9	-0.6	-0.6\\
8	10	-0.9	-0.9\\
9	1	-89.8	-89.8\\
9	2	-60.8	-60.8\\
9	3	-45.9	-45.9\\
9	4	-18.4	-18.4\\
9	5	-9.6	-9.6\\
9	6	-8.7	-8.7\\
9	7	-4.7	-4.7\\
9	8	-1.7	-1.7\\
9	9	-1.1	-1.1\\
9	10	0	0\\
10	1	-104.9	-104.9\\
10	2	-62.9	-62.9\\
10	3	-27.6	-27.6\\
10	4	-18.7	-18.7\\
10	5	-12.5	-12.5\\
10	6	-8.7	-8.7\\
10	7	-6.7	-6.7\\
10	8	-3	-3\\
10	9	-2.2	-2.2\\
10	10	-3	-3\\
};
\end{axis}
\end{tikzpicture}%
} 

\caption{Comparison of the smooth (a) and non-smooth (b) potential. In total $2000$ points are segmented into $4$ clusters with $15$ given sample points for each cluster. With $R=9$ and the number of used eigenvalues at $55$ we obtained $109$ misclassified points in the non-smooth case and $137$ in the smooth case. The difference between the mean for non-smooth (d) and smooth (c) potential is shown in (e). Negative values indicate that the non-smooth potential performed better. \label{fig:datavv1}}%
\end{figure}

We show in Figure \ref{fig:datavv1} the results for a segmentation problem into four classes into the four corners.\footnote{The data are generated using the MATLAB code \url{http://de.mathworks.com/matlabcentral/fileexchange/41459-6-functions-for-generating-artificial-datasets}.} We here vary the number of used eigenvalues of the graph Laplacian as well as the number of samples. We uniformly take the values $n_{sample}=5,10,\ldots,50$ and $k=5,10,\ldots,50$. It can be seen that with an increase in the number of both $n_{sample}$ and $k$ the misclassification is dramatically reduced. Here the one axis shows the variation in $n_{sample}$ and the other the variation in $k$. For the mean we have taken $10$ runs with randomly chosen samples. Figure \ref{fig:datavv1} also shows the difference in the means between the non-smooth and the smooth potential. It can be seen that for sufficient information with larger sample and eigenvalues size the difference is neglectable but for smaller values of $n_{sample}$ the 
non-smooth potential performs better for increasing values of $k$ than the smooth potential. The chosen values are $\omega_0=10000,$ $\nu=10^{-7},$ $\epsilon = 10^{1},$ $\tau = 0.1,$ and $c = (2/\epsilon)+\omega_0.$

\subsection{Hypergraph Laplacian}
We now want to present results for our approach regarding hypergraphs where both the case of a smooth and non-smooth potentials are tested. 
\subsubsection*{Scalar segmentation}
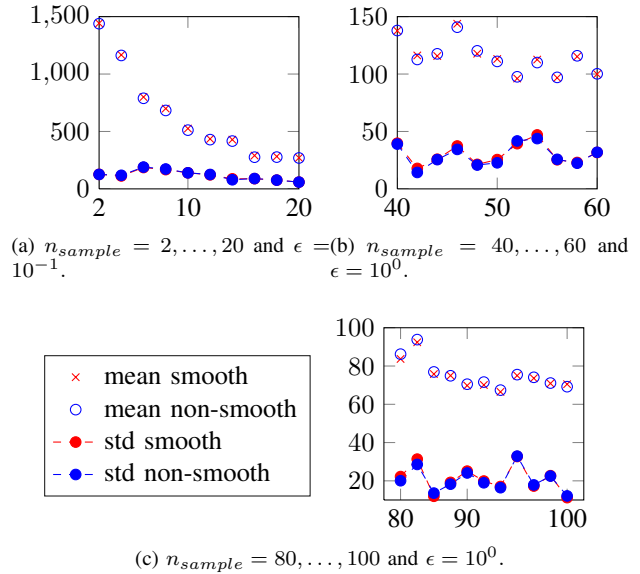
\begin{figure}%
\centering
\subfloat[$n_{sample}=2,\ldots,20$ and $\epsilon=10^{-1}.$]{
%
%
\begin{tikzpicture}
\begin{axis}[%
width=.3\linewidth,
at={(2.684in,1.294in)},
scale only axis,
xtick={1,5,10},
xticklabels={2,10,20},
xmin=1,
xmax=10,
ymin=0,
ymax=1500,
axis background/.style={fill=white},
legend pos=south east
]
\addplot [color=red,only marks,mark=x,mark options={solid},forget plot]
  table[row sep=crcr]{%
1	1442.1\\
2	1166.5\\
3	799.8\\
4	699.4\\
5	525.4\\
6	435.2\\
7	428.8\\
8	290.4\\
9	282.3\\
10	270.7\\
};
\addplot [color=blue,only marks,mark=o,mark options={solid},forget plot]
  table[row sep=crcr]{%
1	1438.5\\
2	1161.9\\
3	788.8\\
4	683\\
5	511.5\\
6	428.8\\
7	416.3\\
8	275.9\\
9	276.7\\
10	269.1\\
};
\addplot [color=red,dashed,mark=*,mark options={solid},forget plot]
  table[row sep=crcr]{%
1	126.274660957771\\
2	112.774332186008\\
3	185.942894459562\\
4	167.620523803024\\
5	139.851492662753\\
6	122.252034747893\\
7	87.2843628607095\\
8	88.8427824868177\\
9	76.0986859282077\\
10	58.9186727616975\\
};
\addplot [color=blue,dashed,mark=*,mark options={solid},forget plot]
  table[row sep=crcr]{%
1	125.800834655419\\
2	118.228126941096\\
3	189.837193405297\\
4	172.994219556608\\
5	139.669073169403\\
6	125.346559585814\\
7	79.2742707314296\\
8	89.950486380008\\
9	76.2194856975564\\
10	58.9414115881186\\
};
\end{axis}
\end{tikzpicture}%
} 
\subfloat[$n_{sample}=40,\ldots,60$ and $\epsilon=10^{0}.$]{
%
%
\begin{tikzpicture}

\begin{axis}[%
width=.3\linewidth,
at={(2.684in,1.294in)},
scale only axis,
xtick={1,6,11},
xticklabels={40,50,60},
xmin=1,
xmax=11,
ymin=0,
ymax=150,
axis background/.style={fill=white}
]
\addplot [color=red,only marks,mark=x,mark options={solid},forget plot]
  table[row sep=crcr]{%
1	137.7\\
2	116.2\\
3	115.7\\
4	143.8\\
5	117.7\\
6	113.1\\
7	96.1\\
8	112.6\\
9	96.7\\
10	115.2\\
11	99.9\\
};
\addplot [color=blue,only marks,mark=o,mark options={solid},forget plot]
  table[row sep=crcr]{%
1	137.9\\
2	112.7\\
3	117.6\\
4	140.8\\
5	120.2\\
6	111\\
7	97.6\\
8	110\\
9	97.1\\
10	115.9\\
11	100.2\\
};
\addplot [color=red,dashed,mark=*,mark options={solid},forget plot]
  table[row sep=crcr]{%
1	39.7543708288787\\
2	17.8202132422707\\
3	25.8149181676022\\
4	37.333095237336\\
5	21.3403373918971\\
6	25.4968625520867\\
7	39.3661021692522\\
8	46.9812728648341\\
9	25.3458083319511\\
10	22.9773801813871\\
11	31.7\\
};
\addplot [color=blue,dashed,mark=*,mark options={solid},forget plot]
  table[row sep=crcr]{%
1	38.9934609902737\\
2	14.3041951888248\\
3	25.4723379374568\\
4	34.3709179394441\\
5	20.7499397589487\\
6	22.702422778197\\
7	41.725771412881\\
8	43.7858424607772\\
9	25.7311095757645\\
10	22.3537468895038\\
11	31.8647140266471\\
};
\end{axis}
\end{tikzpicture}%
} \\

\subfloat[$n_{sample}=80,\ldots,100$ and $\epsilon=10^{0}.$]{
%
%
\begin{tikzpicture}

\begin{axis}[%
width=.3\linewidth,
at={(0.786in,0.498in)},
scale only axis,
xtick={1,5,11},
xticklabels={80,90,100},
xmin=0,
xmax=12,
ymin=10,
ymax=100,
axis background/.style={fill=white},
legend style={at={(-1.7,0.02)},anchor=south west,legend cell align=left,align=left,draw=white!10!black}
]
\addplot [color=red,only marks,mark=x,mark options={solid}]
  table[row sep=crcr]{%
1	83.7\\
2	92.4\\
3	75.8\\
4	75\\
5	69.8\\
6	70.1\\
7	66.5\\
8	75\\
9	73.6\\
10	70.9\\
11	70.5\\
};
\addlegendentry{mean smooth};

\addplot [color=blue,only marks,mark=o,mark options={solid}]
  table[row sep=crcr]{%
1	86.2\\
2	93.8\\
3	76.9\\
4	74.9\\
5	70.5\\
6	71.6\\
7	67.4\\
8	75.5\\
9	74.1\\
10	71.1\\
11	69.2\\
};
\addlegendentry{mean non-smooth};

\addplot [color=red,dashed,mark=*,mark options={solid}]
  table[row sep=crcr]{%
1	22.2263357303897\\
2	31.3534687076247\\
3	11.9398492452794\\
4	19.1989583050748\\
5	25.0790749430676\\
6	19.8919581740964\\
7	17.0308543532026\\
8	32.6894478387139\\
9	17.2812036617824\\
10	22.7\\
11	11.2182886395386\\
};
\addlegendentry{std smooth};

\addplot [color=blue,dashed,mark=*,mark options={solid}]
  table[row sep=crcr]{%
1	20.0638979263751\\
2	28.5720142797108\\
3	13.5974262270475\\
4	18.2671836909798\\
5	24.1215671132702\\
6	18.9905239527507\\
7	16.4146276229466\\
8	32.9127634816647\\
9	17.8686876966385\\
10	22.4831047678029\\
11	12.1227059685534\\
};
\addlegendentry{std non-smooth};

\end{axis}
\end{tikzpicture}%
} 

\caption{Comparison of the misclassification of the smooth (a) and non-smooth (b) potential for the \texttt{mushroom} hypergraph example. We vary the number of sample points and see that for both schemes the results behave similar and the misclassification reduces. \label{fig:data_hg_sclr_1}}%
\end{figure}

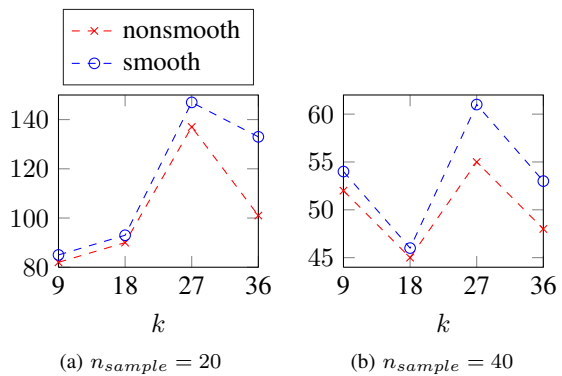
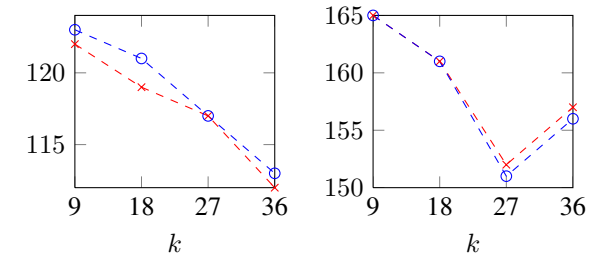
\begin{figure}%
\centering
\subfloat[$n_{sample}=20$]{
%
%
\begin{tikzpicture}

\begin{axis}[%
width=.3\linewidth,
at={(0.786in,0.498in)},
scale only axis,
xtick={1,2,3,4},
xticklabels={9,18,27,36},
xlabel={$k$},
xmin=1,
xmax=4,
ymin=80,
ymax=150,
axis background/.style={fill=white},
legend style={legend cell align=left,align=left,draw=white!15!black,at={(0.5,1.5)},anchor=north}
]
\addplot [color=red,dashed,mark=x,mark options={solid}]
  table[row sep=crcr]{%
1	82\\
2	90\\
3	137\\
4	101\\
};
\addlegendentry{nonsmooth};

\addplot [color=blue,dashed,mark=o,mark options={solid}]
  table[row sep=crcr]{%
1	85\\
2	93\\
3	147\\
4	133\\
};
\addlegendentry{smooth};

\end{axis}
\end{tikzpicture}%
} 
\subfloat[$n_{sample}=40$]{
%
%
\begin{tikzpicture}

\begin{axis}[%
width=.3\linewidth,
at={(2.684in,1.294in)},
scale only axis,
xtick={1,2,3,4},
xticklabels={9,18,27,36},
xlabel={$k$},
xmin=1,
xmax=4,
ymin=44,
ymax=62,
axis background/.style={fill=white}
]
\addplot [color=red,dashed,mark=x,mark options={solid},forget plot]
  table[row sep=crcr]{%
1	52\\
2	45\\
3	55\\
4	48\\
};
\addplot [color=blue,dashed,mark=o,mark options={solid},forget plot]
  table[row sep=crcr]{%
1	54\\
2	46\\
3	61\\
4	53\\
};
\end{axis}
\end{tikzpicture}%
} \\
\subfloat[$n_{sample}=40$ without attribute $32.$]{
%
%
\begin{tikzpicture}

\begin{axis}[%
width=.3\linewidth,
at={(0.786in,0.498in)},
scale only axis,
xtick={1,2,3,4},
xticklabels={9,18,27,36},
xlabel={$k$},
xmin=1,
xmax=4,
ymin=112,
ymax=124,
axis background/.style={fill=white}
]
\addplot [color=red,dashed,mark=x,mark options={solid},forget plot]
  table[row sep=crcr]{%
1	122\\
2	119\\
3	117\\
4	112\\
};
\addplot [color=blue,dashed,mark=o,mark options={solid},forget plot]
  table[row sep=crcr]{%
1	123\\
2	121\\
3	117\\
4	113\\
};
\end{axis}
\end{tikzpicture}%
} 
\subfloat[$n_{sample}=40$ without attribute $32$ and $31.$]{
%
%
\begin{tikzpicture}

\begin{axis}[%
width=.3\linewidth,
at={(0.783in,0.497in)},
scale only axis,
xtick={1,2,3,4},
xticklabels={9,18,27,36},
xlabel={$k$},
xmin=1,
xmax=4,
ymin=150,
ymax=165,
axis background/.style={fill=white}
]
\addplot [color=red,dashed,mark=x,mark options={solid},forget plot]
  table[row sep=crcr]{%
1	165\\
2	161\\
3	152\\
4	157\\
};
\addplot [color=blue,dashed,mark=o,mark options={solid},forget plot]
  table[row sep=crcr]{%
1	165\\
2	161\\
3	151\\
4	156\\
};
\end{axis}
\end{tikzpicture}%
} 


\caption{Comparison of the misclassification of the smooth (a) and non-smooth (b) potential for the \texttt{student performance} hypergraph example. We vary the number of sample points and see that for both schemes the results behave similar and the misclassification reduces. \label{fig:data_hg_school_1}}%
\end{figure}

\begin{figure}%
\centering
\subfloat[Smallest non-zero eigenvectors]{
%
%
\begin{tikzpicture}

\begin{axis}[%
width=.3\linewidth,
at={(0.786in,0.498in)},
scale only axis,
xmin=0,
xmax=100,
ymode=log,
ymin=0.872690340312822,
ymax=0.993331811709706,
yminorticks=true,
axis background/.style={fill=white},
legend style={legend cell align=left,align=left,draw=white!15!black},
legend pos=south west
]
\addplot [color=blue,only marks,mark=o,mark options={solid}]
  table[row sep=crcr]{%
1	0.982078425627293\\
2	0.981609700709081\\
3	0.981416401804682\\
4	0.981121343045258\\
5	0.9808008587549\\
6	0.980506261229494\\
7	0.980036539244145\\
8	0.97999482225222\\
9	0.979573417406552\\
10	0.978941261930643\\
11	0.978666805918689\\
12	0.978427659034336\\
13	0.978030411010276\\
14	0.977509798692396\\
15	0.977445995870412\\
16	0.977066009183157\\
17	0.976563298421134\\
18	0.976266909424772\\
19	0.975809626375524\\
20	0.975050670431158\\
21	0.974951526779536\\
22	0.974006357978437\\
23	0.973912866609878\\
24	0.97370946325436\\
25	0.973384731759434\\
26	0.972737655163359\\
27	0.972308728357289\\
28	0.972190241315209\\
29	0.971846047628669\\
30	0.971439396004438\\
31	0.970850310647291\\
32	0.970181829198737\\
33	0.970112259458301\\
34	0.96974621108366\\
35	0.96971854399552\\
36	0.968909747071991\\
37	0.968243565751642\\
38	0.968071980714778\\
39	0.967578664439143\\
40	0.96696173051941\\
41	0.966579556188033\\
42	0.966489442015719\\
43	0.966123592089004\\
44	0.965509517429171\\
45	0.965416539031481\\
46	0.964899020208978\\
47	0.963707908268952\\
48	0.963374854039333\\
49	0.962969511422793\\
50	0.962612495759264\\
51	0.961866188716878\\
52	0.961285597251582\\
53	0.961179870271252\\
54	0.960960836275775\\
55	0.959953547264298\\
56	0.959016272346697\\
57	0.958836407762167\\
58	0.958113843690751\\
59	0.957630238045278\\
60	0.957345822170049\\
61	0.956957396355971\\
62	0.9561577501823\\
63	0.955733015899819\\
64	0.954922654911187\\
65	0.954422676801378\\
66	0.953542245784412\\
67	0.953402133378224\\
68	0.952573624020899\\
69	0.951434690107772\\
70	0.951275136548729\\
71	0.950423066712714\\
72	0.949406396052705\\
73	0.949171314916212\\
74	0.948091777142543\\
75	0.947171601496314\\
76	0.946022033172655\\
77	0.945877683065584\\
78	0.94502829133033\\
79	0.943509434644544\\
80	0.94141822473622\\
81	0.940455035471796\\
82	0.938314572274858\\
83	0.937902054674165\\
84	0.937609754812892\\
85	0.935243657682901\\
86	0.93436073718358\\
87	0.931618834885477\\
88	0.929449444228599\\
89	0.928192257366007\\
90	0.924909152243172\\
91	0.923968524411253\\
92	0.918529766998254\\
93	0.916199477619989\\
94	0.900290251690309\\
95	0.872690340312822\\
};
\addlegendentry{Hypergraph};

\addplot [color=red,only marks,mark=x,mark options={solid}]
  table[row sep=crcr]{%
1	0.993331811709706\\
2	0.993276453384351\\
3	0.993196210505855\\
4	0.993181461842928\\
5	0.993138760330625\\
6	0.993049351735355\\
7	0.992926547517687\\
8	0.992821419881877\\
9	0.99269607044297\\
10	0.992588228946081\\
11	0.992577250081349\\
12	0.992480773748523\\
13	0.992371164280588\\
14	0.992283288284239\\
15	0.992127863577719\\
16	0.991978351242372\\
17	0.991866196607163\\
18	0.991806837650069\\
19	0.991729198733762\\
20	0.991659169531102\\
21	0.991441774507315\\
22	0.991301247487002\\
23	0.991228010152579\\
24	0.99115755091888\\
25	0.99097629825353\\
26	0.990819432170145\\
27	0.990663116923733\\
28	0.990621111708888\\
29	0.990463987971459\\
30	0.990062080886104\\
31	0.989923394509938\\
32	0.989861663701337\\
33	0.989620229785021\\
34	0.9894290142809\\
35	0.989023735392242\\
36	0.988981267075114\\
37	0.988753121844026\\
38	0.988517936311645\\
39	0.988412022152453\\
40	0.988085121525478\\
41	0.9877825821956\\
42	0.987610577318444\\
43	0.987262731175298\\
44	0.987051369536446\\
45	0.986906106487081\\
46	0.986830878966694\\
47	0.986651842519037\\
48	0.986019668731731\\
49	0.98578858886135\\
50	0.985313236021393\\
51	0.985016831097065\\
52	0.98479650090581\\
53	0.984628013331759\\
54	0.984102444322208\\
55	0.983742734891896\\
56	0.983408506939099\\
57	0.982854804263306\\
58	0.9820950442142\\
59	0.98191029605943\\
60	0.981348937417387\\
61	0.980930172149443\\
62	0.980749136220079\\
63	0.980039085055436\\
64	0.97972255418652\\
65	0.97941491108669\\
66	0.978899354148299\\
67	0.977897091917507\\
68	0.97659686135707\\
69	0.976330963338988\\
70	0.975685719156869\\
71	0.975448057554848\\
72	0.974799840853178\\
73	0.973474337235583\\
74	0.97221680591371\\
75	0.971819879926619\\
76	0.970575565831086\\
77	0.970319363076493\\
78	0.969555463270079\\
79	0.969115173314589\\
80	0.968157494107089\\
81	0.967010512924902\\
82	0.96660023714593\\
83	0.965085012773195\\
84	0.962859079391231\\
85	0.962331457843293\\
86	0.959972590292266\\
87	0.958377812551578\\
88	0.957951985388538\\
89	0.956005549524861\\
90	0.952287291366514\\
91	0.952239553739139\\
92	0.948867843739899\\
93	0.942298831409749\\
94	0.933120792902767\\
95	0.919090935751825\\
};
\addlegendentry{Graph};

\end{axis}
\end{tikzpicture}%
}
\subfloat[Graph-based separation]{
%
%
\begin{tikzpicture}

\begin{axis}[%
width=.3\linewidth,
at={(0.783in,0.497in)},
scale only axis,
xtick={1,2,3,4},
xticklabels={9,18,27,36},
xmin=1,
xmax=4,
ymin=110,
ymax=170,
axis background/.style={fill=white},
legend style={legend cell align=left,align=left,draw=white!15!black}
]
\addplot [color=blue,solid,mark=o,mark options={solid}]
  table[row sep=crcr]{%
1	166\\
2	132\\
3	123\\
4	119\\
};
\addlegendentry{smoth};

\addplot [color=red,solid,mark=x,mark options={solid}]
  table[row sep=crcr]{%
1	166\\
2	132\\
3	122\\
4	117\\
};
\addlegendentry{nonsmooth};

\end{axis}
\end{tikzpicture}%
}
\caption{Comparison of the eigenvalues of the hypergraph Laplacian and the graph Laplacian applied to the school example. The right picture shows the misclassification for the graph Laplacian based segmentation using an increased number of eigenvalues and a sample size $n_{sample}=40.$\label{fig:compeigs}}%
\end{figure}
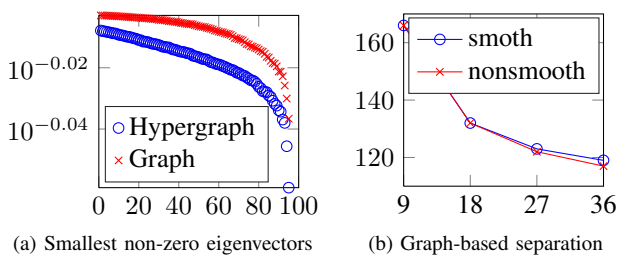

We here focus our attention on two datasets. The first dataset is the so-called \texttt{mushroom} dataset\footnote{We obtain a MATLAB version of the data from \url{http://people.whitman.edu/~hundledr/courses/M350F14/M350/mushrooms.mat}.} as introduced by Schlimmer \cite{schlimmer1987concept,Lichman:2013}. The dataset includes descriptions of hypothetical samples species of mushrooms. The goal is to identify each species as edible or non-edible. The latter includes definitely poisonous, unknown edibility, and not recommended. There is no simple or at least safe rule to determine which class a mushroom belongs to. The dataset we used contains $4062$ mushroom species with $21$ attributes, e.g. one attribute is the cap shape with the attribute values bell, conical, flat, knobbed and sunken. Similar to \cite{ZhoHS06} we create a hyperedge whenever one or more species share the same value of a particular attribute. We simply set the entries in the corresponding column in $H$ to $1$. Based on this adjacency matrix 
and a weight vector with constant weight one we obtain the hypergraph Laplacian $L_s$. For the 
computation of the hypergraph Laplacian\footnote{The MATLAB code is given under \url{http://www.ml.uni-saarland.de/code/hypergraph/hypergraphcut.zip}.} we use the Matlab functions based on \cite{hein2013total}. The results shown in Figure \ref{fig:data_hg_sclr_1} illustrate that our approach utilizing the hypergraph Laplacian allows for a solution to the segmentation problem. The performance both for the smooth and the non-smooth potential gets better with an increasing number of samples. The difference between both is almost neglectable even though the non-smooth potential gives slightly better results for small sample sizes but at a higher cost due to the nonlinear iteration at its core. The parameters for both methods are chosen as
$\omega_0 = 10^{5},$ $\tau = 0.1$ $c = (3/\epsilon)+\omega_0,$ and $\nu=1e-3.$
The second example is also taken from the UCI machine learning repository \cite{Lichman:2013} and is the so-called \texttt{student performance} data set as introduced in \cite{cortez2008using}. The data is given for $395$ students with attributes ranging from family size to the job of the parents. All in all $30$ attributes are given with three additional columns noting the grades for the first period, the second period, and the final grade. We follow the approach given in \cite{cortez2008using} by adding these information to the basis on which the hypergraph is formed. Again for all $30$ attributes one or more pupils share a hyperedge whenever they share an attribute value. Additionally, we include hyperedges for the pupils with the same grades based on the first and/or second period. We always run $5$ tests for each scenario and show the mean in Figure \ref{fig:data_hg_school_1}. The parameters for this example are given via
$\omega_0 = 10^8,$ $\epsilon = 10^{-2},$ $\tau = 0.1,$ $c = (3/\epsilon)+\omega_0$ and $\nu=10^{-6}$.
We also show the difference in the eigenvalues of the hypergraph Laplacian and the graph Laplacian using a weight matrix $W$. In order to generate the matrix $W$ we take the feature vector for each of the $395$ pupils and use \eqref{weightgraph}. In Figure \ref{fig:compeigs} we show the difference in the smallest non-zero eigenvalues of the two Laplacians as well as the separation for the school example when the graph Laplacian is used. The parameters are set to $\omega_0=10^8,$ $\epsilon = 10^{-2},$ $\tau = 0.1,$ $c = (3/\epsilon)+\omega_0$ and $\nu=10^{-6}.$ It can be seen that the segmentation improves with an increasing number of eigenvectors and we note that we have chosen the same parameters as for the hypergraph Laplacian. It is not clear whether this parameter constellation is the best possible as in this setup the hypergraph Laplacian outperforms the graph Laplacian.

\subsubsection*{Multiclass segmentation}
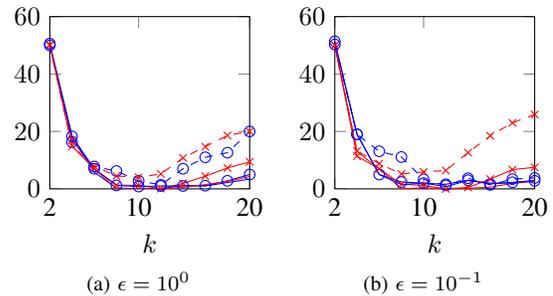
\begin{figure}%
\centering
\subfloat[$\epsilon=10^{0}$]{
%
%
\begin{tikzpicture}

\begin{axis}[%
width=.3\linewidth,
at={(2.684in,1.243in)},
scale only axis,
xtick={1,5,10},
xticklabels={2,10,20},
xlabel={$k$},
xmin=1,
xmax=10,
ymin=0,
ymax=60,
axis background/.style={fill=white},
legend style={legend cell align=left,align=left,draw=white!15!black}
]
\addplot [color=blue,dashed,mark=o,mark options={solid}]
  table[row sep=crcr]{%
1	50\\
2	16.3\\
3	7.8\\
4	6.2\\
5	2.8\\
6	1.4\\
7	7\\
8	11\\
9	12.6\\
10	20\\
};

\addplot [color=red,dashed,mark=x,mark options={solid}]
  table[row sep=crcr]{%
1	50\\
2	14.9\\
3	7.9\\
4	4.5\\
5	4\\
6	5.2\\
7	10.8\\
8	14.7\\
9	18.6\\
10	20\\
};

\addplot [color=red,solid,mark=x,mark options={solid}]
  table[row sep=crcr]{%
1	50.2\\
2	17.1\\
3	7\\
4	1.3\\
5	1.2\\
6	0.1\\
7	1.9\\
8	4.4\\
9	7.3\\
10	9.4\\
};

\addplot [color=blue,solid,mark=o,mark options={solid}]
  table[row sep=crcr]{%
1	50.6\\
2	18.3\\
3	7\\
4	1.2\\
5	0.8\\
6	1\\
7	1\\
8	1.1\\
9	2.8\\
10	5\\
};

\addplot [color=blue,solid]
  table[row sep=crcr]{%
1	50\\
2	18.4\\
3	5.8\\
4	0.3\\
5	0\\
6	0\\
7	1.2\\
8	1\\
9	2.2\\
10	3.4\\
};

\addplot [color=red,solid]
  table[row sep=crcr]{%
1	50\\
2	14.9\\
3	7\\
4	0\\
5	0\\
6	0\\
7	0.5\\
8	1.5\\
9	2.6\\
10	4.3\\
};

\end{axis}
\end{tikzpicture}%
} 
\subfloat[$\epsilon=10^{-1}$]{
%
%
\begin{tikzpicture}

\begin{axis}[%
width=.3\linewidth,
at={(2.684in,1.243in)},
scale only axis,
xtick={1,5,10},
xticklabels={2,10,20},
xlabel={$k$},
xmin=1,
xmax=10,
ymin=0,
ymax=60,
axis background/.style={fill=white},
legend style={legend cell align=left,align=left,draw=white!15!black}
]
\addplot [color=blue,dashed,mark=o,mark options={solid}]
  table[row sep=crcr]{%
1	50.3\\
2	19\\
3	13.1\\
4	11.1\\
5	3.3\\
6	1.2\\
7	3.7\\
8	1.4\\
9	3.4\\
10	3.8\\
};

\addplot [color=red,dashed,mark=x,mark options={solid}]
  table[row sep=crcr]{%
1	50\\
2	13.2\\
3	8.8\\
4	5.2\\
5	5.9\\
6	6.4\\
7	12.6\\
8	18.5\\
9	23\\
10	25.9\\
};

\addplot [color=blue,solid,mark=o,mark options={solid}]
  table[row sep=crcr]{%
1	51.4\\
2	19\\
3	5\\
4	2.4\\
5	2\\
6	1.6\\
7	3\\
8	1.7\\
9	2.3\\
10	2.8\\
};

\addplot [color=red,solid,mark=x,mark options={solid}]
  table[row sep=crcr]{%
1	50\\
2	11.4\\
3	7\\
4	1.6\\
5	1.2\\
6	0\\
7	0.5\\
8	3.4\\
9	6.7\\
10	7.5\\
};

\addplot [color=red,solid]
  table[row sep=crcr]{%
1	50\\
2	13\\
3	7\\
4	0\\
5	0.4\\
6	0\\
7	0.2\\
8	0.6\\
9	1.9\\
10	2.6\\
};

\addplot [color=blue,solid]
  table[row sep=crcr]{%
1	51\\
2	19\\
3	5\\
4	1.5\\
5	2\\
6	0.8\\
7	2.9\\
8	2\\
9	2\\
10	3\\
};

\end{axis}
\end{tikzpicture}%
} 

\caption{Comparison of the misclassification of the mean of $10$ runs of the smooth vs. non-smooth potential. Blue lines represent the non-smooth and the red ones the smooth potentials. Dashed lines correspond to one given sample point, solid lines with markers to two given sample points, and solid lines alone to three given sample points per class. Shown is the total misclassification against the number of eigenvectors used. The left plot is for $\eps=10^{0}$ and the right one for $\eps=10^{-1}$. \label{fig::hg_vv}}
\end{figure}

We again use an example from the UCI ML repository. In particular, we focus on the \texttt{zoo} dataset introduced in \cite{forsyth1990pc}. This dataset contains $101$ individuals with $18$ attributes such number of legs or whether they have hair. The segmentation is performed into $7$ classes that are already pre-specified. We want our algorithm to segment the data into these $7$ classes given only a small number of samples from each class. Figure \ref{fig::hg_vv} shows the results for a small number of samples for each class as well as a varying number of eigenvectors of the hypergraph Laplacian. We also test two different values of the interface parameter $\epsilon$. The results for the non-smooth potential tend to be slightly better than for the smooth potential, especially when the number of eigenvectors grows. We have set the parameters to  $\omega_0=100$ $\epsilon = 10^{-1},$ $\tau = 0.01,$ $c= (3/\epsilon)+\omega_0,$ and $\nu=10^{-4}$ in this example.
\section*{Conclusions and outlook}
We have shown that diffuse interface methods while already being very powerful can be further generalized. We illustrated that non-smooth potentials are a viable option for the separation of data. While the computations become more expensive due to the nonlinearity that is treated with the semi-smooth Newton scheme, the results in many cases show that the results are even better than for the smooth potential. Additionally, we showed that the methods are not limited to the graph Laplacian setup but can successfully be employed for the hypergraph Laplacian. Future work should incorporate more sophisticated eigenvalue methods and our goal is to further investigate different techniques for the segmentation of hypergraphs.
\section*{Acknowledgment}
The third author would like to thank Cristina Garcia-Cardona for answering questions regarding her work on diffuse interface methods on graphs.

\ifCLASSOPTIONcaptionsoff
  \newpage
\fi



%

\bibliographystyle{IEEEtran} 
\bibliography{refs,data,publications}

\end{document}